\documentclass[sigconf, screen]{acmart}
\AtBeginDocument{%
    \hypersetup{
        colorlinks=true,
        allcolors=black,
        citecolor=blue
    }%
  
}

\copyrightyear{2026}
\acmYear{2026}
\setcopyright{cc}
\setcctype{by}
\acmConference[MM '26]{Proceedings of the 34th ACM International Conference on Multimedia}{November 10--14, 2026}{Rio de Janeiro, Brazil}
\acmBooktitle{Proceedings of the 34th ACM International Conference on Multimedia (MM '26), November 10--14, 2026, Rio de Janeiro, Brazil}
\acmDOI{10.1145/3767308.3836170}
\acmISBN{979-8-4007-2213-4/2026/11}

\newcommand{\method}{\textsc{MarsTSC}}
\newcommand{\partitle}[1]{\smallskip \noindent \textbf{#1.}}

\usepackage{makecell,bm}
\usepackage[table]{xcolor}
\usepackage{array}
\usepackage[most]{tcolorbox}
\usepackage{algorithm}
\usepackage{algorithmic}
\usepackage{amsmath} 
\usepackage{enumitem}
\usepackage{ulem}
\usepackage{placeins}
\usepackage{float}
\usepackage{booktabs}
\usepackage{graphicx}
\usepackage{multirow}
\usepackage{balance}
\usepackage{xcolor,colortbl}
\usepackage{enumitem}
\usepackage{longtable}
\usepackage{tcolorbox}
\usepackage{wrapfig}

\tcbuselibrary{skins, breakable}
\definecolor{BrownRed}{RGB}{180, 60, 50}
\definecolor{PrimaryBlue}{RGB}{0, 105, 180}

\newcommand{\repourl}[1]{%
    \begingroup
    \hypersetup{urlcolor=magenta}%
    \def\UrlFont{\ttfamily}%
    \url{#1}%
    \endgroup
}

\begin{document}

\title{Empowering VLMs for Few-Shot Multimodal Time Series Classification via Tailored Agentic Reasoning}

\author{Lin Li}
\authornote{Equal contribution.}
\email{lilin236@mail2.sysu.edu.cn}
\orcid{0009-0005-3755-1329}
\affiliation{%
  \institution{Sun Yat-sen University}
  \city{Zhuhai}
  \state{Guangdong}
  \country{China}
}

\author{Jiawei Huang}
\authornotemark[1]
\email{huangjw255@mail2.sysu.edu.cn}
\orcid{0009-0009-9542-1273}
\affiliation{%
  \institution{Sun Yat-sen University}
  \city{Zhuhai}
  \state{Guangdong}
  \country{China}
}

\author{Qihao Quan}
\email{quanqh@mail2.sysu.edu.cn}
\orcid{0009-0008-6957-9088}
\affiliation{%
  \institution{Sun Yat-sen University}
  \city{Zhuhai}
  \state{Guangdong}
  \country{China}
}

\author{Dan Li}
\authornote{Corresponding author.}
\email{lidan263@mail.sysu.edu.cn}
\orcid{0000-0002-3787-1673}
\affiliation{%
  \institution{Sun Yat-sen University}
  \city{Zhuhai}
  \state{Guangdong}
  \country{China}
}

\author{Boxin Li}
\email{liboxin1@xiaomi.com}
\orcid{0009-0003-6152-764X}
\affiliation{%
  \institution{Xiaomi Corporation}
  \city{Beijing}
  \country{China}
}

\author{Xiao Zhang}
\email{zhangxiao16@xiaomi.com}
\orcid{0009-0004-6423-5914}
\affiliation{%
  \institution{Xiaomi Corporation}
  \city{Beijing}
  \country{China}
}

\author{Erli Meng}
\email{mengerli@xiaomi.com}
\orcid{0009-0009-3748-126X}
\affiliation{%
  \institution{Xiaomi Corporation}
  \city{Beijing}
  \country{China}
}

\author{Wenjie Feng}
\email{fengwenjie@ustc.edu.cn}
\orcid{0000-0003-3636-0035}
\affiliation{%
  \institution{University of Science and Technology of China}
  \city{Hefei}
  \state{Anhui}
  \country{China}
}

\author{Jian Lou}
\email{louj5@mail.sysu.edu.cn}
\orcid{0000-0002-4110-2068}
\affiliation{%
  \institution{Sun Yat-sen University}
  \city{Zhuhai}
  \state{Guangdong}
  \country{China}
}

\author{See-Kiong Ng}
\email{seekiong@nus.edu.sg}
\orcid{0000-0001-6565-7511}
\affiliation{%
  \institution{National University of Singapore}
  \city{Singapore}
  \country{Singapore}
}


\renewcommand{\shortauthors}{Lin Li et al.}

\begin{abstract}
As a rich fusion of multiple types of media, time series encompass correlated and complementary representations across numeric, textual, and visual formats, and are pervasive across diverse real-world application domains. Harnessing recent advances in large vision-language models (VLMs) offers a promising direction for exploiting the rich intrinsic temporal characteristics embedded across these complementary media for few-shot time series classification.  However, existing approaches rely on static context that fail to automatically evolve with the few-shot training samples, and lack tailored agentic reasoning mechanisms to maximally squeeze out discriminative knowledge from the limited labeled data.

In this paper, we propose the first VL\underline{\textbf{M}} \underline{\textbf{a}}gentic \underline{\textbf{r}}easoning framework for few-\underline{\textbf{s}}hot multimodal \underline{\textbf{T}}ime \underline{\textbf{S}}eries \underline{\textbf{C}}lassification (\method{}), which introduces a self-evolving knowledge bank as a dynamic context iteratively refined via reflective agentic reasoning. The framework comprises three collaborative roles: i) Generator conducts reliable classification via reasoning; ii) Reflector diagnoses the root causes of reasoning errors to yield discriminative insights targeting the temporal features overlooked by Generator; iii) Modifier applies verified updates to the knowledge bank to prevent context collapse. We further introduce a test-time update strategy to enable cautious, continuous knowledge bank refinement to mitigate few-shot bias and distribution shift. Extensive experiments across 12 mainstream time series benchmark datasets demonstrate that \method{} delivers substantial and consistent performance gains across 5 VLM backbones, outperforming both classical and foundation model-based time series baselines under few-shot conditions, while producing interpretable rationales that ground each classification decision in human-readable feature evidence. Code is available at \repourl{https://github.com/HuangJW0821/MarsTSC}.
\end{abstract}




\maketitle

\section{Introduction}
Time series classification (TSC) is a fundamental task prevalently encountered across diverse real-world domains, including healthcare monitoring~\citep{morid2023time, yang2025diagecg}, industrial pattern recognition~\citep{abburi2023closer, farahani2025time}, environmental sensing~\citep{du2019deep, mulayim2026can}, and financial analysis~\citep{ang2025structured,chan2004time}, where accurately capturing temporal patterns and precisely identifying contextually relevant information greatly impacts decision-making quality. While traditional TSC methods rely solely on numeric sequence representations~\cite{middlehurst2021hivecote, tan2022multirocket}, the recent rise of large vision-language models (VLMs) \cite{chen2024internvl, liu2023llava, yang2025qwen3} unlocks the opportunity to jointly process numeric, textual, and visual time series representations \citep{ni2025harnessing, zhong2025timevlm}. These multimodal time series constitute correlated yet complementary sources of information across media, offering substantially greater potential for accurate and interpretable classification. Emerging work has begun to leverage VLMs for multimodal time series in tasks such as forecasting \cite{jiang2026timexl}, retrieval \cite{chen2025trace}, and causal discovery  \cite{li2024realtcd}, exploiting their rich intrinsic knowledge acquired during extensive pretraining. Moreover, the inherent reasoning capability of VLMs supports interpretable decision making via chain-of-thought trajectories that provide grounded rationales, which can be particularly valuable in high-stakes application scenarios such as industrial fault detection~\citep{khaniki2024enhanced}, financial analysis~\citep{calik2025explainable}, and medical diagnosis~\citep{moin2024exploring}. These promising results collectively suggest that VLMs hold significant promise for advancing multimodal time series classification.

Despite this potential, the application of VLMs to multimodal time series classification remains largely underexplored, particularly in the challenging few-shot setting where only a limited number of labeled samples are available per class \cite{jiang2025multisurvey, barreda2024cosco}. Performance in this setting hinges critically on two factors: how to maximally capture information from the limited labeled samples and pass it to the VLM, and how to construct the proper context to induce the VLM's intrinsic knowledge and reasoning capabilities. Context is especially consequential for time series data, since the same time series values can carry very different meanings depending on the accompanying contextual information, which is unlike natural text, images, or audio where semantics are relatively stable across contexts. Straightforward approaches include fine-tuning VLM parameters on the limited training samples, or employing in-context learning to prompt the VLM directly \cite{zhao2025images, wang2025tabletime, xue2023promptcast}. Existing attempts have also explored summarizing statistics from the numeric modality, feeding textual and visual time series to VLMs, and designing reasoning templates tailored for time series discovering that such tailored templates often yield better performance than unguided VLM reasoning \citep{zhou2025enhancing}. However, these approaches suffer from two fundamental limitations. First, they lack the capacity to automatically evolve the context so that the VLM can make the best use of the limited labeled samples, adapt its prediction behavior to the classification task, and mitigate few-shot bias. Second, existing methods lack a mechanism to precisely ground their decision-making process, i.e., both intermediate rationales and final predictions, to specific evidence in the multimodal time series data, limiting the ability to refine the context in a targeted and precise manner. This raises a key research question: 

\textit{How can we guide VLMs' reasoning behavior to construct a proper and dynamic context for multimodal time series classification in the few-shot setting?}

In this paper, we answer this question by proposing \method{}, a VL\underline{\textbf{M}} \underline{\textbf{a}}gentic \underline{\textbf{r}}easoning framework for few-\underline{\textbf{s}}hot multimodal \underline{\textbf{T}}ime \underline{\textbf{S}}eries \underline{\textbf{C}}lassification. During the training stage, rather than updating VLM model parameters or relying on static context, \method{} introduces a dynamic knowledge bank that is iteratively optimized through reflective agentic reasoning over the few-shot training samples, serving as an evolving context for classification. During the testing stage, equipped with the knowledge bank that maintains discriminative class characteristics, \method{} produces predictions and reasoning trajectories by grounding its decision-making process in the precise evidence locatable in the knowledge bank, while cautiously updating it to avoid few-shot bias. Both stages are non-trivial. For knowledge bank maintenance and refinement, solely relying on the VLM's built-in reasoning capability risks context collapse, i.e., either over-rewriting existing content or incorrectly modifying the wrong knowledge entries during iterative refinement. For test-time adaptation, the few-shot setting inevitably introduces bias, and natural distribution shift during inference makes it non-trivial to enable reflective, cautious self-updating without compounding errors.

To overcome these challenges, \method{} devises the knowledge bank to contain structured feature bullet points and enable atomic, incremental updates with restricted actions (ADD, MODIFY, DELETE) applied on a per-bullet basis, thereby increasing the precision with which individual feature entries can be located and updated. \method{} proposes the agentic reasoning framework comprising three collaborative roles: the \textbf{Generator}, which performs classification by grounding its predictions in established knowledge; the \textbf{Reflector}, which diagnoses classification outcomes to extract discriminative temporal patterns and error-avoidance strategies from the reasoning trajectory; and the \textbf{Modifier}, which addresses few-shot bias and context collapse by applying verified incremental updates to the knowledge bank. Extensive experiments on 12 benchmark datasets across 7 domains corroborate the superiority of \method{} over both classical TSC methods, VLM/LLM-based methods, and time series foundation models in the few-shot setting. It also achieves substantial performance gains across 5 different base models, demonstrating strong transferability. Moreover, \method{} achieves favorable performance even when compared against methods trained under full-shot scenarios with at most 22 times more training samples. 

In summary, our key contributions are as follows.

\begin{itemize}[leftmargin=*]
    \item[$\bullet$] We study the under-explored task of multimodal time series classification in the few-shot setting, proposing a tailored agentic reasoning framework to elicit the full potential of VLMs across complementary numeric, textual, and visual TS representations.
    \item[$\bullet$] We propose \method, a framework of time-series-tailored agentic reasoning for few-shot multimodal time series classification. It consists of three key stages: initializing task-related knowledge from visual patterns and statistical features, iteratively extracting reasoning insights and inference strategies through agentic refinement, and cautiously evolving the knowledge bank during inference via a test-time update mechanism.
    \item[$\bullet$] We conduct extensive experiments demonstrating that \method{} delivers substantial and consistent performance gains across 5 VLM backbones, outperforming both classical and foundation model-based time series baselines under few-shot conditions.
\end{itemize}

\begin{figure*}[!t]
    \centering 
    \includegraphics[width=0.85\textwidth]{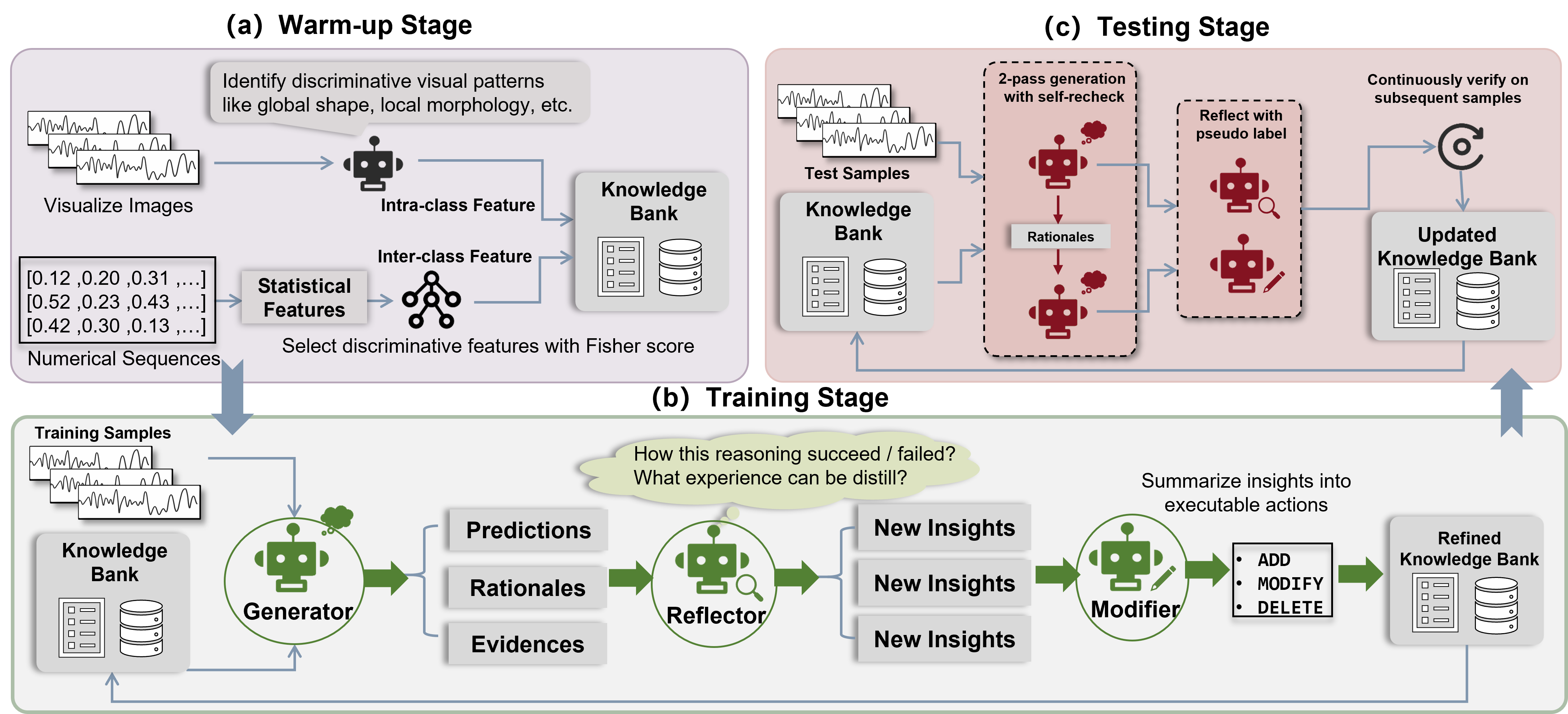} 
    \caption{Overview of our \method{} framework. It comprises three key stages: \textit{Warm-up Stage}, \textit{Training Stage} and \textit{Testing Stage}.}
    \vspace{-1em}
\label{fig:overview} 
\end{figure*}

\section{Related Work}

\partitle{Time Series Classification}
Time series classification (TSC), as a foundational data analytics task, is drawing extensive research interest across numerous domains. Traditionally, TSC methods were dominated by machine learning techniques trained with numerical TS data. Approaches like K-NN \citep{cover1967knn}, decision tree \citep{quinlan1986decision}, and SVM \citep{cortes1995svm} showed satisfactory results via an in-domain learning paradigm. As the demand for higher performance and robustness grew, ensemble methods like random forest \citep{breiman2001randomforest} and HIVE-COTE \cite{middlehurst2021hivecote} achieved state-of-the-art performance by aggregating diverse feature representations and heterogeneous classifiers. 

The paradigm subsequently shifted with the rise of deep learning, which bypassed manual feature engineering in favor of automatic representation learning. Foundational architectures such as RNN \cite{mohammadi2024deep} and Transformer \cite{vaswani2017attention} model sequential dependencies through recurrent state transitions and attention mechanisms, respectively, which naturally led to more sophisticated designs, including TimesNet \cite{wu2023timesnet}, MultiRocket \cite{tan2022multirocket}, Autoformer \cite{wu2021autoformer}, and PatchTST \cite{nie2023patchtst}.

Recent progress in LLMs has fueled increasing interest in adopting large models for time series tasks \citep{jin2023large, guan2026timeomni, hu2026mindts, liu2025llad, zhou2023one}. Early attempts include adapting pretrained LLMs by fine-tuning on abundant datasets using textualized or patched representations of raw time series. For instance, Time-LLM \citep{jin2023timellm} reprograms time series into textualized prompts, LLM4TS \citep{chang2025llm4ts} aligns patched sequences with LLMs via two-stage fine-tuning. However, these efforts mostly focus on improving task performance via modality alignment between textual descriptions and numerical time series \citep{ge2025eventtsf, liu2025timecma, lei2026mmists, wang2026unlocking, chen2025cctime}, while neglecting the reasoning capabilities of large models and the agent paradigm. 

On the other hand, time series foundation models (TSFM) pre-trained with large-scale cross-domain numerical time series datasets emerge as a family of promising generalized time series analysis methods \cite{goswami2024moment, ansari2024chronos, liu2025moirai, ning2025tsrag, wu2026aurora}. Although the aforementioned foundation models show satisfactory classification results, most of them cannot handle multimodal data and lack the reasoning ability to explicitly explain the thinking process.

Time-series data are also commonly represented through textual and visual modalities other than numerical \cite{wu2026rating, shen2026multi}. Driven by the success of works like LLaVA~\citep{liu2023llava} and Qwen2.5-VL~\citep{bai2025qwen25vltechnicalreport}, researchers have shown increasing interest in leveraging Vision Language Mod-\\els (VLMs) for time series analysis, unlocking a nascent yet promising frontier for addressing the aforementioned limitations of existing time series approaches~\citep{zhao2025images}. Preliminary attempts suggest that VLMs could capture important temporal patterns such as trends, periodicity, and spikes during their extensive pre-training on massive imaged time series from various domains~\citep{ni2025harnessing, daswani2024plots, zhuang2024seeit}. Furthermore, compared with LLM-based time series analysis relying on discrete textual descriptions~\citep{tan2024language}, VLMs are claimed to be able to leverage imaged time series data in compact and continuous representations, which can retain more temporal information and be less restricted by the context length~\citep{daswani2024plots}. 

\partitle{Agents and Self-Improvement Techniques for Reasoning Tasks}
The evolution of Large Language Models (LLMs) has catalyzed the development of autonomous agents capable of complex reasoning, tool manipulation, and sequential decision-making. Pioneering frameworks such as ReAct \cite{yao2023react} interleave reasoning and acting, allowing models to dynamically adjust their execution plans based on environmental feedback, and TRT \cite{zhuang2026testtime} proposed an iterative self-improvement framework through self-generated verification signals. To maintain context over extended tasks, incorporating structured memory has become essential for agentic frameworks. For instance, architectures like A-mem \cite{xu2025amem} and ACE \cite{zhang2025ace} utilized learned memory for code and math tasks. 

Recently, these agentic paradigms have been extended into the domain of time series reasoning \citep{zhou2026time, ning2025towards, ruan2026visual, wu2026timeart}. For example, TimeXL \cite{jiang2026timexl} integrates a prototype-based time series encoder with three collaborating LLMs to deliver more accurate predictions and interpretable explanations; MERIT \cite{zhou2025merit} utilizes three LLM agents to collaboratively generate positive views for time series data; and TSci \cite{zhao2025timeseriesscientist} performs data analysis and forecasting by integrating specialized agents.
However, these paradigms lack exploration in the reasoning ability of agentic VLMs, especially tailored for the interpretable TSC task that requires a detailed understanding of hidden time series patterns. 

\section{Problem Formulation}
Time-series classification aims to learn a mapping from time-series data to class labels. Formally, let $D = \{(x_i, y_i)\}_{i=1}^{N}$ denote a time series dataset with $N$ samples, where $x_i \in \mathbb{R}^{m \times w}$ is a multivariate time-series sample with $m$ variables and $w$ time steps, and $y_i \in \{1, 2, ..., c\}$ denotes the corresponding label among $C$ classes. The dataset is partitioned into training, validation, and test splits, denoted by $D_{\text{train}}$, $D_{\text{val}}$, $D_{\text{test}}$, respectively. In particular, this paper focuses on the few-shot setting where the sample size of training and validation sets is very small per class (e.g., 3 samples), representing a realistic yet more challenging scenario as considered in previous work \cite{vinyals2016matching}. Unlike full-shot scenarios where models can be trained from scratch or sufficiently fine-tuned with ample labeled samples, few-shot classification demands maximally exploiting the very limited training samples, posing a fundamental challenge for capturing and refining class-discriminative knowledge from scarce examples.

During the training stage, instead of optimizing model parameters on $D_{\text{train}}$, our agentic multimodal time series classification framework $\mathcal{F}$ introduces an optimizable knowledge bank $\mathcal{K}$ that serves as dynamic contextual knowledge, updated through an iterative agentic reasoning framework. Through iterative refinement by the proposed agentic reasoning framework, $\mathcal{K}$ is gradually enriched with task-related experience and insights during training, and continuously updated during testing. 

During the testing stage, the \method{} framework $\mathcal{F}$ classifies a given test sample $x$ as:
    $\hat{y} = \mathcal{F}(x, \mathcal{K}, \mathcal{C})$,
where $\mathcal{K}$ denotes the refined knowledge bank, and $\mathcal{C}$ represents auxiliary contextual knowledge, including the dataset background and sample-level statistical characteristics. Together, $\mathcal{K}$ and $\mathcal{C}$ provide the agentic context for the classification decision-making process. 

\section{Methodology}
\subsection{Overview of Our Proposed \method{}}
We present an overview of the proposed \method{} for multimodal few-shot time series classification, organized into three stages: Warm-up, Training, and Testing. The overall framework of \method{} is illustrated in Figure~\ref{fig:overview}. 

\noindent\textbf{Warm-up Stage.} \method{} sets up the knowledge bank by extracting prior knowledge from few-shot training samples, including core statistical features (e.g., mean value, min/max value, frequency) from numerical time series and visual features (e.g., trend, periodicity, spike) from line plots. This warm-up step equips $\mathcal{F}$ with essential domain-specific priors, allowing the agents to anchor their general pre-trained knowledge to the specific temporal patterns of the task. However, this time series knowledge remains static and suboptimal, lacking the discriminative refinement necessary for accurate classification. While previous methods have largely relied on such static contexts alone, our \method{} treats them only as a stepping stone for the subsequent iterative refinement during the training and testing stages, as introduced below.

\noindent\textbf{Training Stage.} Our \method{} optimizes the knowledge bank on the few-shot training and validation samples to gradually refine the task contexts in a dynamic manner through tailored iterative reasoning, rather than training or fine-tuning model parameters, which are known to be less effective in the few-shot setting. \method{} organizes the agentic reasoning framework into three roles, i.e., Generator, Reflector, and Modifier, which cooperate to refine the knowledge bank according to the trial-and-error feedback obtained from decision-making on training and validation samples. 

\noindent\textbf{Testing Stage.} \method{} allows agents to continuously refine the knowledge bank at test-time, so as to further increase the dynamism of the agent's task-related knowledge, thereby mitigating distribution shift during inference. Specifically, we introduce a robust Test-Time Update mechanism designed to selectively refine the knowledge bank. By leveraging self-recheck through two-pass generation, pseudo-label-based refinement and gated knowledge management, our method is able to cautiously yet effectively adapt to previously unseen temporal patterns and extract test-time task-related reasoning strategies in an unsupervised manner.

\subsection{Warmup by Atomic Knowledge Bank Design}

\begin{figure}[h]
    \centering 
    \vspace{-1.0em}
    \includegraphics[width=0.85\columnwidth]{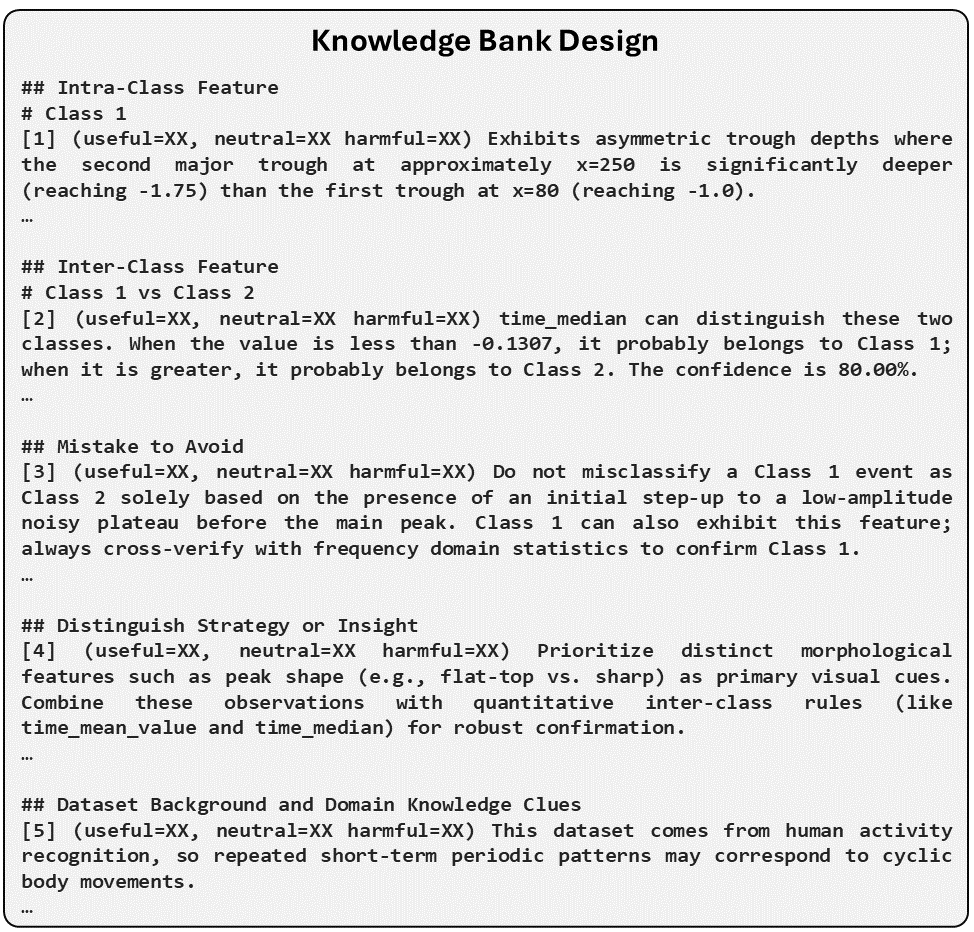} 
    \vspace{-1.25em}
    \caption{Illustration of knowledge bank design.}
    \vspace{-1.5em}
\label{fig:knowledge_base_example} 
\end{figure}

\partitle{Knowledge Bank Design}
The key design choice of \method{} lies in introducing a dynamic knowledge bank that maximally squeezes contextual information out from the very few training and validation samples and optimizes the knowledge rather than updating model parameters to support subsequent classification. The structure of the knowledge bank is pivotal, as solely relying on the VLM's built-in reasoning capability risks over-updating the current knowledge by completely rewriting its existing content or incorrectly updating the wrong knowledge entries.  Therefore, we structure the knowledge bank as feature bullet points to enable atomic updates, thereby increasing the precision with which individual feature entries can be located. Each bullet is maintained by a unique ID and a tag counter (useful, neutral, useless). Specifically, Figure~\ref{fig:knowledge_base_example} provides an illustration of the knowledge bank design, which contains the following sections: (1) Intra-class features; (2) Inter-class features; (3) Mistakes to avoid; (4) Distinguishing strategies or insights; (5) Dataset background and domain knowledge clues.

\partitle{Intra-class Feature Initialization} 
We construct intra-class feature descriptors to capture prototypical characteristics of samples for each class, covering both the numeric and imaged time series modalities and organizing their complementary features jointly. Specifically, we leverage the cross-modal alignment capabilities of VLMs to analyze the visual representations of the imaged time series modality and extract intra-class semantic descriptors, which are then combined with statistical and domain knowledge extracted from the numeric time series modality.

\partitle{Inter-class Feature Initialization} 
Inter-class feature descriptors capture the most salient distinguishing characteristics between pairs of classes. The aim is to equip the knowledge bank with explicit discriminative boundaries that guide the agent in differentiating visually or semantically similar classes during classification. However, extracting such boundaries is non-trivial: while VLMs excel at capturing semantic and visual patterns, they lack the quantitative precision required for determining fine-grained classification boundaries, necessitating a complementary statistical analysis beyond semantic initialization alone. We compute a normalized Fisher Score to evaluate pairwise class separability, establishing explicit decision thresholds for the most discriminative descriptors between each pair of classes. After filtering out redundant metrics, only the most critical statistical descriptors are retained and incorporated into the inter-class feature entries of the knowledge bank.

\subsection{Training by Iterative Reflective Reasoning}

To dynamically refine the knowledge bank, thereby implicitly improving the precision of locating the decision boundaries between classes, we introduce an iterative reflective reasoning framework comprising three collaborative roles: a Generator, a Reflector, and a Modifier. Rather than updating model parameters, these three roles engage in a cooperative trial-and-error process over the few-shot training and validation samples. In detail, the Generator produces classification decisions and rationales that the Reflector diagnoses, and the Reflector's extracted lessons are then structured and committed to the knowledge bank by the Modifier. This cooperative loop allows the framework to backtrack from suboptimal feature bullets, avoid compounding errors across reasoning rounds, and iteratively converge toward a more accurate and generalized knowledge bank through accumulated trial-and-error experience.

\partitle{Generator}
The Generator serves as the entry point of each reflective reasoning round, producing a classification decision explicitly grounded in the current state of the knowledge bank. 
Given a query time-series sample $x_{q}$ and the current knowledge bank $\mathcal{K}_t$, the Generator is prompted to refer to $\mathcal{K}_t$ for decision-making evidence and cites the specific feature bullet points $\mathcal{T}$ from $\mathcal{K}_t$ that directly support its classification prediction. This grounded reasoning produces a predicted class label $\hat{y}$ alongside a chain-of-thought reasoning trajectory $\mathcal{R}$ as the rationale that makes the knowledge bank usage explicit and traceable:  
\begin{equation}
    (\hat{y}, \mathcal{R}, \mathcal{T}) = \text{Generator}(x_{q}, \mathcal{K}_t, \mathcal{C}).
\end{equation}
\partitle{Reflector}
The Reflector serves as the evaluation component of each reasoning round, diagnosing the Generator's decision by comparing the predicted class $\hat{y}$ against the ground truth $y_{gt}$ and identifying the precise causes of any reasoning errors. Upon evaluating the prediction, the Reflector tags the cited feature bullet points as either \textit{useful}, \textit{neutral}, or \textit{useless} based on their contribution to the correct or incorrect decision. 
When a misclassification occurs ($\hat{y} \neq y_{gt}$), the Reflector retrieves representative hard-positive samples $x^{+}$ from the ground-truth class $y_{gt}$ and hard-negative samples $x^{-}$ from the incorrectly predicted class $\hat{y}$. By contrasting the differences among $x_{q}$, $x^{+}$, and $x^{-}$, it identifies the root causes of the original reasoning error and yields concrete and discriminative insights $\mathcal{I}_{new}$ that isolate the exact temporal features the Generator overlooked or underestimated. These contrastive samples serve as explicit comparative anchors that guide the Reflector in pinpointing fine-grained discriminative boundaries overlooked by the Generator. 

\partitle{Modifier}
We find that naive self-improvement that directly prompts the model to refine the knowledge bank leads to overwriting the entire content. As the knowledge bank grows, monolithic rewriting inevitably causes context degradation, where information is suddenly erased or compressed, hampering the convergence of the iterative refinement process and yielding degraded or irrelevant feature descriptions. To ensure atomic and precise refinement, the Modifier is restricted to performing updates bullet by bullet. It reads the Reflector's new insights $\mathcal{I}_{new}$, reviews the existing knowledge bank, and outputs structured, granular actions $action \in \{\text{ADD}, \text{MODIFY}, \text{DELETE}\}$. Instead of regenerating the entire repository, it produces a compact set of incremental updates that are applied directly to the knowledge bank in a structured and automated manner. This strategy allows the knowledge bank to incrementally accumulate new insights:
\begin{equation}
    \mathcal{K}_{t+1} = \mathcal{K}_t \oplus \text{Modifier}(\mathcal{I}_{new}, \mathcal{K}_t, \mathcal{C}).
\end{equation}
\partitle{Discarding}
Real-world time series data is notoriously susceptible to transient anomalies, sensor artifacts, and atypical outliers. Spurious rules pollute the knowledge bank, causing a noise-induced bias that severely degrades generalization on subsequent samples. To safeguard the knowledge bank against such pollution, \method{} incorporates a discarding mechanism driven by the Reflector's tagging history. This is crucial because models may generate unsupported details in their rationales \citep{ji2023hallucination, huang2025survey}. To mitigate this, we maintain a dynamic counter for each bullet $k_i \in \mathcal{K}$. If the number of useless tags exceeds a threshold $\tau_{\text{drop}}$, or if the ratio of useful to useless tags indicates that the bullet is ambiguous, the Modifier forcefully purges or refines it.
This strategy reduces spurious noise and ensures that only statistically recurrently supported temporal patterns are permanently consolidated into the knowledge bank.

\subsection{Testing with Evolving Knowledge Bank}
The knowledge extracted from few-shot samples is highly dependent on the quality of those samples, and may therefore lack sufficient generalization and transferability. This limitation becomes even more pronounced in time series classification, where distribution shifts are common and further expose the shortcomings of a static knowledge bank. To address this, we propose continuously evolving the knowledge bank at test-time to dynamically capture distributional patterns absent from the few-shot training samples. However, the test stage is inherently unsupervised, and naively treating model predictions as reliable supervision risks knowledge corruption that undermines downstream classification performance. To this end, we propose a Test-Time Update mechanism that enables cautious and reliable knowledge bank evolution during inference.

\partitle{Two-pass Generation}
To reduce reasoning bias and mitigate hallucinations, we propose a two-pass generation mechanism. Specifically, the Generator first makes a standard inference for the current test sample, producing the first-pass predicted label $\hat{y}_1$ and corresponding reasoning $\mathcal{R}_1$:
\begin{equation}
    (\hat{y}_1, \mathcal{R}_1, \mathcal{T}_1) = \text{Generator}(x, \mathcal{K}, \mathcal{C}).
\end{equation}
In the second pass, $(\hat{y}_1, \mathcal{R}_1)$ is integrated into the Generator's context and will be treated as a suspicious reference. The generator is then prompted to re-examine whether the prior reasoning was overconfident or overlooked important evidence, and to produce a new prediction:
\begin{equation}
    (\hat{y}_2, \mathcal{R}_2, \mathcal{T}_2) = \text{Generator}(x, \mathcal{K}, \mathcal{C}, \hat{y}_1, \mathcal{R}_1).
\end{equation}
The second-pass prediction is taken as the final answer. This mechanism helps better exploit the self-correction ability of large models~\citep{pang2023language, bensal2025reflect, tie2025correctbench}, and improves reasoning accuracy through self-recheck.

\partitle{Refine on Pseudo Label}
Another purpose of our two-pass generation design is to provide a pseudo-supervisory signal for test-time refinement. Specifically, we treat second-pass prediction as a pseudo-label and perform reflection under the hypothesis that the current test sample belongs to class $\hat{y}_2$. We first examine the consistency between the two passes. If the two predictions disagree, we regard this as an indication that $\hat{y}_1$-related knowledge is unreliable or insufficient. In this case, the test sample is treated as a negative instance and $\hat{y}_1$-related knowledge is sharpened accordingly. In contrast, if the two predictions are consistent, the test sample is treated as a positive instance for class $\hat{y}_2$ and the associated knowledge is further refined.

The refinement process is carried out by the Reflector and Modifier using the same prompts as in the training stage. The only difference is that the Modifier is not allowed to perform the \text{DELETE} operation, in order to accommodate the update design introduced in the following section.

\partitle{Deferred Knowledge Update}
When the two-pass generation mechanism produces the same incorrect prediction in both passes, the system may still associate the test sample with an incorrect label, and the risk of knowledge contamination remains. To address this issue, we propose a deferred knowledge update mechanism, which continuously validates newly acquired knowledge during the test stage.

Specifically, we partition the knowledge memory into three levels according to confidence: prototype knowledge, reference knowledge, and candidate knowledge.

\vspace{-0.3em}
\begin{itemize}[leftmargin=*]
\item \textit{Prototype knowledge} refers to the knowledge acquired during warm-up and training stages. Since it is generated with labeled samples, it is treated as reliable ground-truth and serves as the knowledge bank foundation. Prototype knowledge is frozen during testing and the Modifier is not allowed to operate on it, thereby preventing contamination of the core knowledge.

\item \textit{Reference knowledge} refers to test-time knowledge that has been repeatedly observed in subsequent samples. It is provided to the Generator as a suspicious but informative reference.

\item \textit{Candidate knowledge} refers to newly extracted knowledge at test-time. It is temporarily stored in a buffer and does not participate in classification decisions.
\end{itemize}

To support knowledge evolution, we maintain a counter for each candidate and reference item. After the modifier completes its operation, we check whether the candidate knowledge associated with $\hat{y}_2$ is observed in the current sample. Its counter is increased by 1 if it is observed, otherwise decreased by 1. When the score exceeds threshold $\tau_{\text{promote}}$, the candidate is promoted to reference knowledge; when it falls below threshold $\tau_{\text{remove}}$, it is then removed from the knowledge memory. For reference knowledge, we further adjust its counter according to the Reflector’s \textit{useful/useless} judgment. If the score drops below $\tau_{\text{promote}}$, it is demoted back to candidate knowledge. Through this deferred update design, the system continuously filters and calibrates test-time knowledge, ensuring that only patterns and insights repeatedly supported by subsequent observations are incorporated into decision-making, thereby further improving the quality of the knowledge bank.

\vspace{-0.25em}
\section{Experiments}
\label{sec:exps}
\vspace{-0.25em}

\subsection{Experimental Settings}
\vspace{-0.3em}

\partitle{Dataset}
We conduct experiments on 12 datasets collected from the UCR/UEA archive~\citep{dau2019ucr, bagnall2018uea}, covering diverse scenarios with varying numbers of variables and classes. For each dataset, we follow the default train--test split, and further construct few-shot train and validation sets by randomly sampling from the original training split. Specifically, we set $k=3$ by default, meaning that 3 examples are sampled for each class at most, in both $D_{\text{train}}$ and $D_{\text{val}}$.

\partitle{Baselines}
The proposed method is compared against several representative TSC baselines, spanning into three categories: \textit{machine learning-based methods} including DrCIF, HIVE-COTE 2.0~\citep{middlehurst2021hivecote}, \textit{deep learning-based methods} including TimesNet~\citep{wu2023timesnet}, Autoformer~\citep{wu2021autoformer}, PatchTST~\citep{nie2023patchtst}, and \textit{large model-based methods} including GPT4TS~\citep{zhou2023one}, MOMENT~\citep{goswami2024moment}.

In accordance with the few-shot training manner of our proposed method, all baselines are evaluated under similar few-shot settings in the main results. Meanwhile, since most of the baselines are not designed as few-shot TSC methods, for a fair and comprehensive comparison, we also compare the proposed method (few-shot) with baselines under the full-shot setting, where they are trained or fine-tuned on the original full training split. 

\partitle{Implementation Details}
Line plot is adopted as visual representation, as it explicitly presents the shape-based temporal characteristics. We use Gemini-3.1-pro~\citep{gemini31pro_modelcard} as the base VLM of \method{}. Results with other VLMs are shown in the appendix. For test-time update, test samples are processed sequentially in a randomly permuted order, and knowledge bank is reset before each run.

\partitle{Metric}
We use classification accuracy as the evaluation metric. All results are averaged over three independent runs with different random seeds.

\partitle{Other Details}
Additional details, ablations, and analysis are provided in Appendix \ref{sec:app_implementation} and \ref{sec:more_analysis}.

\begin{table*}[!t]
\centering
\caption{Classification accuracy under few-shot setting. Best results are in \textbf{Bold}, and second-best results are \underline{Underlined}.}
\label{tab:result_few_shot}
\vspace{-0.1in}
\setlength{\tabcolsep}{4pt}
\renewcommand{\arraystretch}{0.5}
\scriptsize
\resizebox{\textwidth}{!}{
\begin{tabular}{l|c c c c c c c c}
\toprule
Dataset & \method{} & DrCIF & HIVE-COTE 2.0 & TimesNet & Autoformer & PatchTST & GPT4TS & MOMENT \\
\midrule
ArrowHead & \textbf{0.766} & 0.554  & \underline{0.669} & 0.606 & 0.451 & 0.303 & 0.411 & 0.606 \\
DodgerLoopGame & \textbf{0.884} & 0.739 & \underline{0.746} & 0.681 & 0.580 & 0.522 & 0.529 & 0.710 \\
ECG200 & \textbf{0.790} & 0.660 & \underline{0.770} & 0.660 & 0.690 & 0.760 & 0.760 & 0.750 \\
GesturePebbleZ1 & \textbf{0.785} & 0.262 & 0.209 & \underline{0.302} & 0.157 & 0.209 & 0.168 & 0.198 \\
Lightning7 & \textbf{0.822} & 0.658 & 0.671 & 0.616 & 0.603 & 0.315 & 0.452 & \underline{0.726} \\
UMD & \textbf{0.993} & \underline{0.799} & 0.625 & 0.549 & 0.514 & 0.507 & 0.583 & 0.757 \\
Worms & \textbf{0.481} & 0.260 & 0.442 & 0.182 & 0.208 & 0.286 & 0.260 & \underline{0.442} \\
DiatomSizeReduction & \underline{0.971} & 0.964 & \textbf{0.980} & 0.804 & 0.843 & 0.284 & 0.882 & 0.768 \\
DistalPhalanxTW & \underline{0.576} & 0.568 & 0.554 & 0.468 & 0.511 & 0.144 & \textbf{0.612} & 0.511 \\
Fish & \underline{0.771} & 0.743 & \textbf{0.849} & 0.509 & 0.509 & 0.183 & 0.286 & 0.709 \\
RacketSports & \underline{0.526} & 0.513 & \textbf{0.579} & 0.401 & 0.263 & 0.250 & 0.467 & 0.520 \\
SelfRegulationSCP1 & \underline{0.697} & 0.659 & \textbf{0.791} & 0.677 & 0.536 & 0.529 & 0.642 & 0.689 \\
\midrule
\rowcolor{cyan!8} \textbf{1st Count} & \textbf{7} & 0 & \underline{4} & 0 & 0 & 0 & 1 & 0 \\
\bottomrule
\end{tabular}
}
\end{table*}

\begin{table}[t]
\centering
\caption{Accuracy comparison with direct inference.}
\label{tab:ablation_reasoning}
\vspace{-0.1in}
\small
\renewcommand{\arraystretch}{0.8}
\begin{tabular}{l|ccc}
\toprule
Dataset & Ours & base+text+num &base+text+plot \\
\midrule
DodgerLoopGame & \textbf{0.884} & 0.607 & 0.623 \\
DiatomSizeReduction & \textbf{0.971} & 0.255 & 0.291 \\
DistalPhalanxTW & \textbf{0.576} & 0.180 & 0.173 \\
Fish & \textbf{0.771} & 0.143 & 0.137 \\
GesturePebbleZ1 & \textbf{0.785} & 0.164 & 0.146  \\
\bottomrule
\end{tabular}
\end{table}

\begin{table}[!h]
\centering
\vspace{-0.8em}
\caption{Ablation on test-time update.}
\label{tab:ablation_test_time_update}
\vspace{-0.125in}
\small
\renewcommand{\arraystretch}{0.8}
\begin{tabular}{l|cccccc}
\toprule
Dataset & Full pipeline & w/o update & w/o 2-pass \\
\midrule
DodgerLoopGame & \textbf{0.884} & 0.812 & 0.659 \\
DiatomSizeReduction & \textbf{0.971} & 0.791 & 0.782 \\
DistalPhalanxTW & \textbf{0.576} & 0.561 & 0.482 \\
Fish & \textbf{0.771} & 0.731 & 0.686 \\
GesturePebbleZ1 & \textbf{0.785} & 0.767 & 0.773  \\
\bottomrule
\end{tabular}
\vspace{-1.7em}
\end{table}

\partitle{Research Questions}
To comprehensively evaluate the effectiveness of the proposed framework, we conduct experiments to answer the following four Research Questions (RQs).

\noindent\textbf{RQ1:} Can the proposed agentic TS reasoning framework outperform traditional machine learning and deep learning methods, and existing large model-based methods under a few-shot setting? 

\noindent\textbf{RQ2:} Can the performance of the proposed agentic TS reasoning framework under fixed few-shot setting compete with that of baselines under increasing $k$-shot settings (until full-shot settings)?

\noindent\textbf{RQ3:} Can the proposed agentic TS reasoning framework with reflection and test-time update yield performance gains compared to direct LLM/VLM inference?

\noindent\textbf{RQ4:} Can the agentic rechecking and updating pipeline during the testing stage indeed improve the quality of the knowledge bank and enhance the overall performance of \method{}?

\vspace{-0.75em}
\subsection{Main Results}

\subsubsection{Comparison under Few-shot Setting (RQ1)}
\label{main_comparison}
The proposed method is compared with baselines under the same few-shot setting ($k$=3, 3 for training and another 3 for validation). As shown in Table~\ref{tab:result_few_shot}, our method consistently achieves the best or second-best performance across all benchmarks. These results highlight the key advantage of our proposed method that it can achieve strong few-shot classification performance by leveraging multimodal information and agentic reasoning rather than relying on end-to-end parameter learning with large-scale samples.

\vspace{-0.25em}
\subsubsection{Comparison with Increasing K-shot Baselines (RQ2)}
\begin{figure}[b]
    \centering
    \vspace{-1.25em}
    \includegraphics[width=0.85\linewidth]{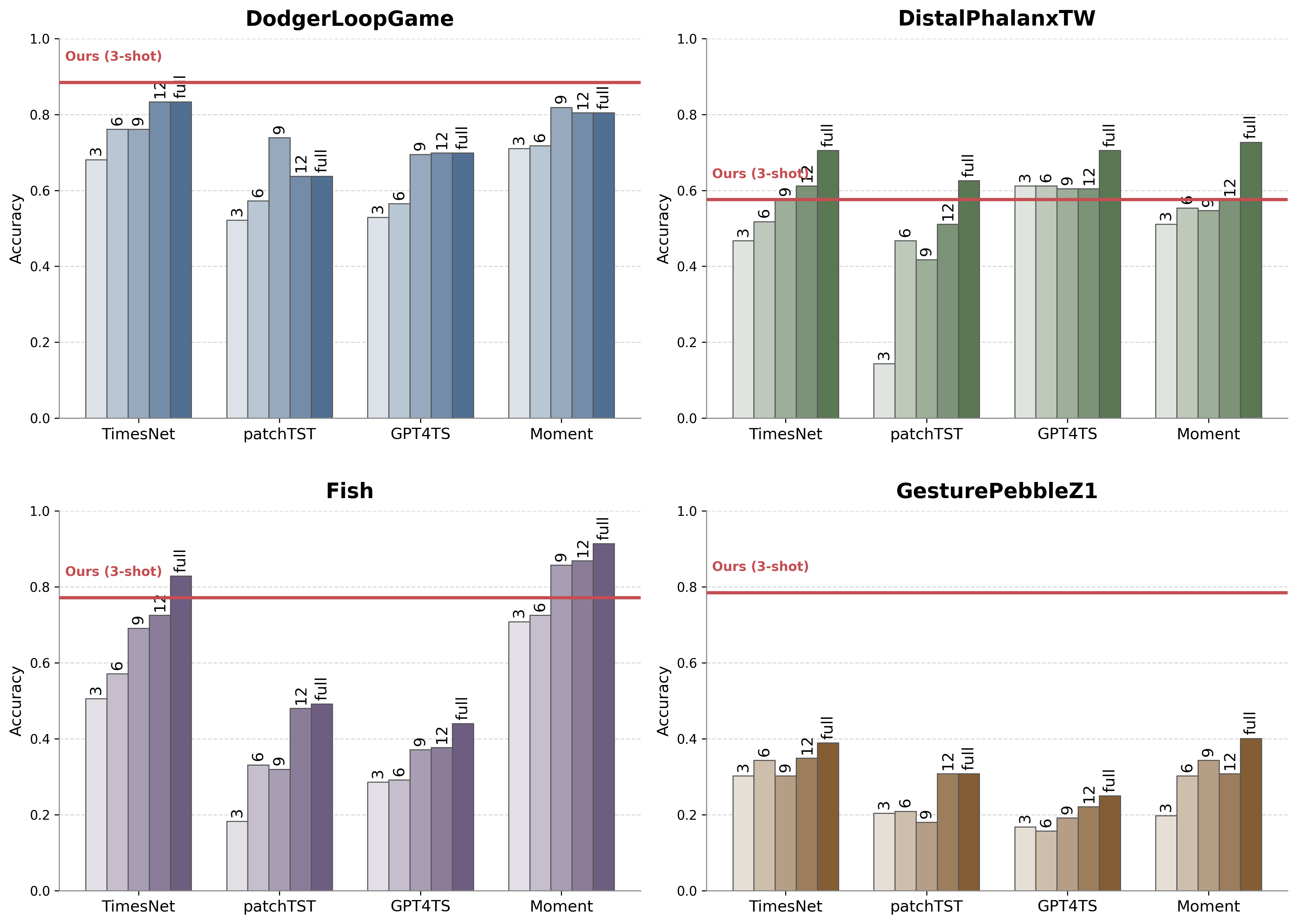} 
    \vspace{-1em}
    \caption{Comparison with Increasing $k$-shot Baselines }
    \vspace{-1.5em}
\label{fig:comparison} 
\end{figure}

In this comparison, our proposed method remains fixed in the 3-shot setting, whereas the baselines gradually increase the $k$ values until full-shot. The results, as shown in Figure~\ref{fig:comparison}, indicate that our method remains competitive with, and in most cases even outperforms, baselines trained with substantially more data. This suggests that, with a proper reasoning procedure, few-shot samples can be exploited much more effectively, thereby unlocking stronger performance. 

\subsubsection{Effects of TS-tailored Agentic Reasoning (RQ3)}
In this section, we compare the proposed \method{} with a direct inference setting, where the base model (Gemini-3.1-pro, marked as ``base'' in the table) is prompted to conduct TS classification with merely dataset descriptions (marked as ``text'' in the table) and numerical time-series (marked as ``num'' in the table) or visualized time-series plots (marked as ``plot'' in the table), without any agentic reflections. This experiment aims to measure how much improvement can be brought by the proposed time-series tailored agentic TS reasoning framework. 

Results in Table \ref{tab:ablation_reasoning} show that directly applying the VLM base model with plain textual descriptions and TS plots yields limited performance and consistently underperforms our method, which is equipped with tailored agentic TS reasoning. This suggests that generic visual-language reasoning capabilities alone are insufficient to reliably solve time-series classification, especially in the few-shot setting. The clear advantage of our method indicates that the real performance gain does not come merely from using a stronger foundation model, but from introducing a more suitable task-adaptation reasoning paradigm for time-series classification.

\begin{figure}[t]
    \centering 
    \includegraphics[width=0.7\linewidth]{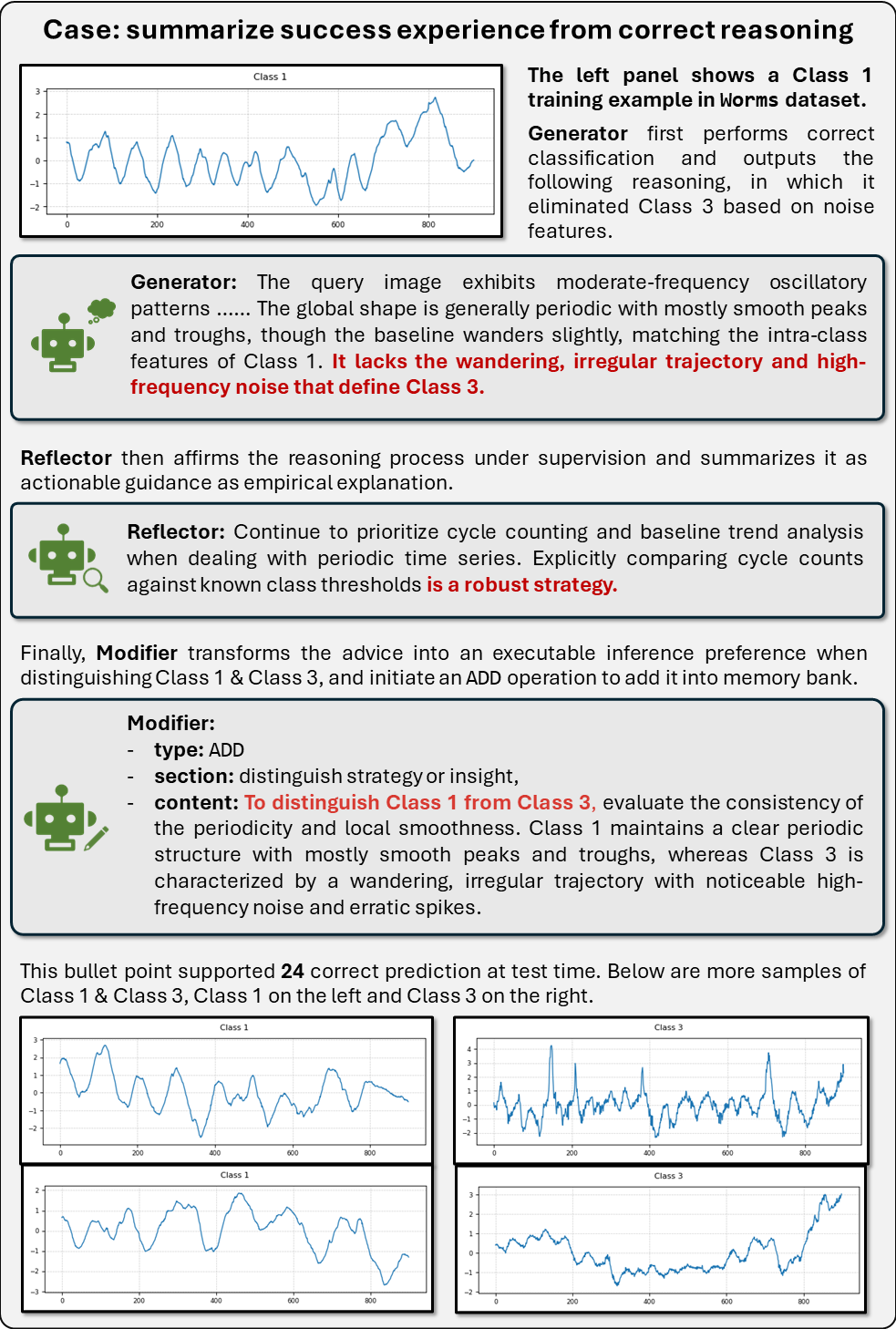} 
    \vspace{-1.35em}
    \caption{A case about train refinement on success reasoning. Through refinement, the agent summarizes successful experiences into reusable and executable guidelines.}
    \vspace{-1.85em}
\label{fig:case_1} 
\end{figure}

\vspace{-0.85em}
\subsection{Ablation Studies on Test Update (RQ4)}

To answer RQ4, we conduct ablation studies to show the effectiveness of the test update pipeline.

As shown in Table~\ref{tab:ablation_test_time_update}, the ablation variant \textit{w/o update} retains the two-pass generation for self-recheck in the testing stage but disables refine and update on knowledge bank. Its inferior performance shows that the feature descriptions derived from few-shot samples alone cannot fully cover the diversity of test examples, while test-time update helps capture additional patterns beyond the limited training data. 

The ablation variant  \textit{w/o 2-pass} removes the self-recheck step and directly uses the first-pass reasoning for refinement and update, resulting in a more substantial performance drop. This suggests that the self-recheck process enables the model to correct unreliable initial reasoning before updating the feature bank. Without this step, erroneous knowledge from the mistake reasoning process is more likely to be incorporated, leading to knowledge contamination and degraded classification performance. More ablations about the key reasoning steps are in the Appendix \ref{ab_on_key_stages}.

\subsection{Case Studies of Rechecking and Test-Time Update (RQ4)}
\begin{figure}[t]
    \centering 
    \includegraphics[width=0.70\linewidth]{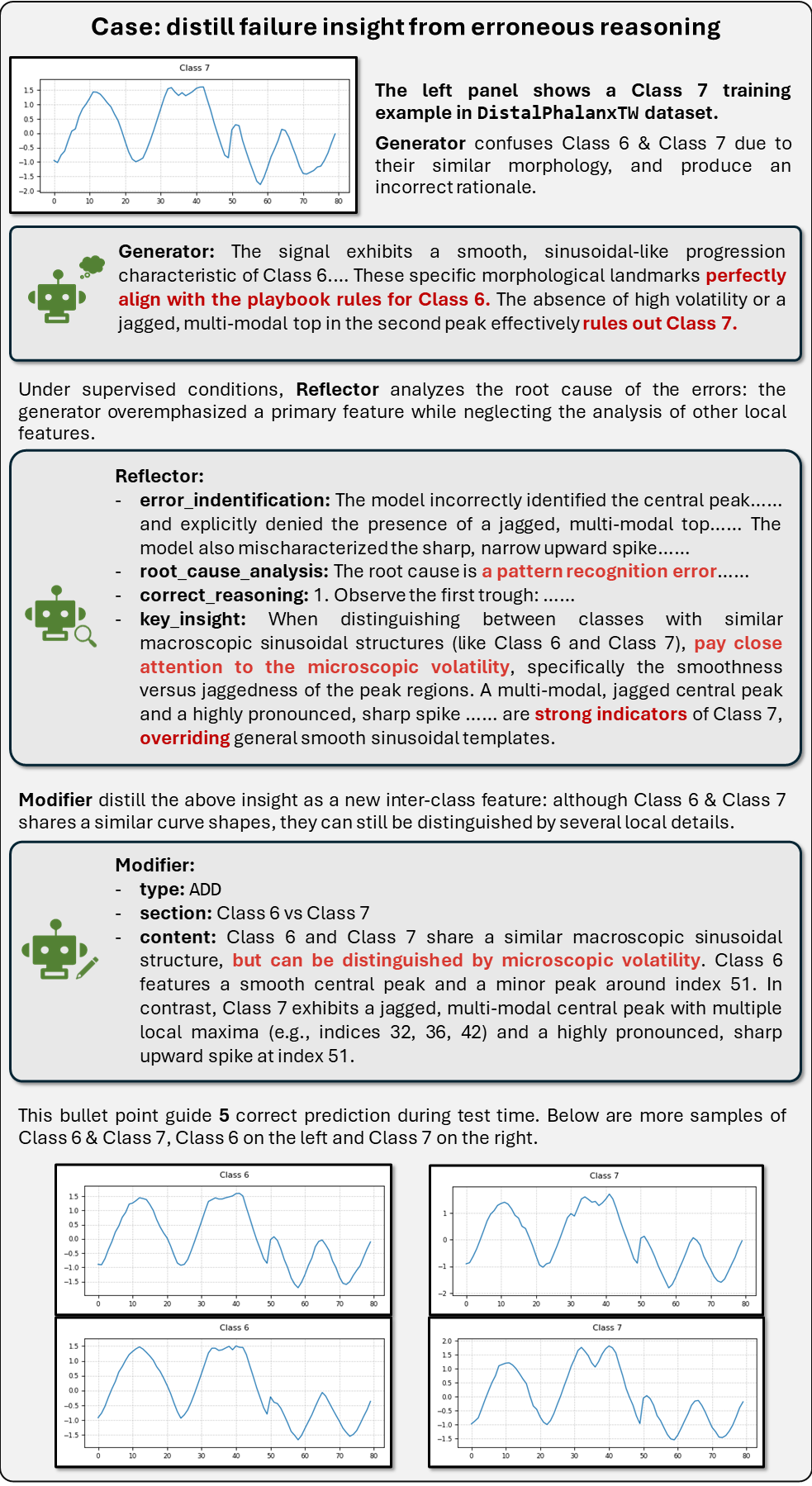} 
    \vspace{-1.25em}
    \caption{A case about train refinement on erroneous reasoning. Through refinement, the agent learns more accurate task knowledge.}
    \vspace{-2em}
\label{fig:case_2} 
\end{figure}

This section illustrates the effectiveness of the designed rechecking and updating processes through detailed case studies. Figure~\ref{fig:case_1} shows train-time refinement from a correct prediction output, where the reflector summarizes successful experience from the reasoning process and the modifier further distills it into an actionable distinguishing strategy. Figure~\ref{fig:case_2} shows train-time refinement from an incorrect prediction output, where the reflector inspects the reasoning and identifies the root cause of the error upon an incorrect prediction, and then derives a corrected reasoning process and summarizes the resulting failure insight. These two cases demonstrate that the reflection process at training time could explicitly summarize successful experiences and identify error causes to improve the quality of the maintained knowledge bank.

We also present a case of a successful test-time self-recheck via two-pass generation, where the model revisits its previous reasoning, corrects the initial mistake, and ultimately arrives at the correct answer. This case suggests that the test-time rechecking and updating reasoning manner can effectively reduce the risk of error-induced knowledge contamination while further expanding the coverage and adaptability of the knowledge bank. Details can be found in Appendix \ref{sec:more_case_studies}.

\vspace{-0.5em}
\section{Conclusion}

In this paper, we have proposed \method{}, the first VLM agentic reasoning framework for few-shot multimodal time series classification. Unlike existing approaches that rely on static context and suboptimally exploit limited labeled samples, \method{} introduces a self-evolving knowledge bank iteratively refined through three collaborative roles, i.e., a Generator, a Reflector, and a Modifier, alongside a Test-Time Update strategy that mitigates few-shot bias during inference. Extensive experiments demonstrate superior few-shot classification accuracy over both classical and VLM-based baselines, with interpretable rationales grounded in human-interpretable feature evidence.

\begin{acks}
The authors thank the anonymous reviewers and area chairs for their constructive comments and efforts in improving this paper. 

We also gratefully acknowledge the National Supercomputer Center in Guangzhou for the computational resources and support provided.
\end{acks}

\bibliographystyle{ACM-Reference-Format}
\bibliography{imagedTS}

\clearpage
\appendix
\section{Appendix Overview}
This appendix is organized as follows: We first describe additional implementation details and prompting settings in Appendix~\ref{sec:app_implementation}. We then provide supplementary experiments in Appendix \ref{sec:more_analysis}, including extended ablations and more case studies. Finally, Appendix \ref{sec:future_directions} outlines several promising directions for future research.

\section{Additional Implementation Details}
\label{sec:app_implementation}
\partitle{Dataset Selection}
Rather than pursuing exhaustive coverage of the entire UCR/UEA archive, we evaluate \method{} on 12 representative datasets spanning diverse domains and temporal characteristics. For more detailed comparisons and ablations, we further adopt a fixed diagnostic subset, which is used consistently across further analysis to ensure comparability and reduce experiment-specific selection bias.

\begin{table*}[t]
\centering
\normalsize
\caption{Details of datasets}
\label{tab:datasets}
\vspace{-1em}
\renewcommand{\arraystretch}{0.9}
\begin{tabular}{lcccccc}
\toprule
Dataset & Train size & Test size & Classes & Length & Variables & Type \\
\midrule
ArrowHead  & 36 & 175 & 3 & 251 & 1 & IMAGE \\
DodgerLoopGame  & 20 & 138 & 2 & 288 & 1 & SENSOR \\
ECG200  & 100 & 100 & 2 & 96 & 1 & ECG \\
GesturePebbleZ1  & 132 & 172 & 6 & N/A & 1 & HAR \\
Lightning7  & 70 & 73 & 7 & 319 & 1 & SENSOR \\
UMD  & 36 & 144 & 3 & 150 & 1 & SIMULATED \\
Worms  & 181 & 77 & 5 & 900 & 1 & MOTION \\
DiatomSizeReduction  & 16 & 306 & 4 & 345 & 1 & IMAGE \\
DistalPhalanxTW  & 400 & 139 & 6 & 80 & 1 & IMAGE \\
Fish  & 175 & 175 & 7 & 463 & 1 & IMAGE \\
RacketSports  & 151 & 152 & 4 & 30 & 6 & HAR \\
SelfRegulationSCP1 & 268 & 293 & 2 & 896 & 4 & EEG \\

\bottomrule
\end{tabular}
\end{table*}

\partitle{Hyperparameter Details}
During training, we set the discarding threshold $\tau_{\text{drop}}$ to 10\% of the total training steps, scaling it proportionally with the size of the train set. During testing, by contrast, we heuristically set the promotion threshold $\tau_{\text{promote}}$ to 5 and the removal threshold $\tau_{\text{remove}}$ to -2, so as to simulate a real-world deployment scenario where test samples arrive sequentially in a streaming manner.

\partitle{Visualization Details}
Line plot is adopted as visual representation. For multivariate datasets, each variate is rendered as an individual subplot with channel index. The original numerical scale is preserved.

\partitle{VLMs Details}
We evaluated \method{} using 5 representative base model from 4 model families: Gemini~\citep{gemini31pro_modelcard}, GPT~\citep{openai_gpt5, openai_gpt54mini}, Kimi~\citep{kimi_k25}, and Qwen~\citep{qwen35_397b}. This selection spans diverse parameter scales and is sourced from different corporate developers. Detailed information about these VLMs is summarized in Table \ref{tab:vlms_overview}.

For the main experiments, Gemini serves as the primary base model, while Appendix~\ref{ab_on_vlm} reports the results obtained when utilizing the other models. For all experiments, the temperature is fixed at 0.2.

\begin{table*}[t]
\centering
\normalsize
\caption{Details of adopted base VLMs}
\label{tab:vlms_overview}
\vspace{-1em}
\renewcommand{\arraystretch}{0.9}
\begin{tabular}{lccccc}
\toprule
VLM & Type & Parameters & Developer & Release Date \\
\midrule
GPT-5.4 mini  & Proprietary & N/A  & OpenAI & 2026-03 \\
GPT-5  & Proprietary & N/A  & OpenAI  & 2025-08 \\
Gemini-3.1 pro  & Proprietary & N/A   & Google & 2026-02 \\
Kimi-K2.5 & Open-source & 1000B total / 32B activate & Moonshot AI & 2026-01 \\
Qwen3.5-397B-A17B & Open-source & 397B total / 17B activate  & Alibaba & 2026-02 \\
\bottomrule
\end{tabular}
\end{table*}

\partitle{Prompt Details} In this section, we provide our prompts for agents in \method{}.

\definecolor{BrownRed}{RGB}{180, 60, 50}
\definecolor{PrimaryBlue}{RGB}{154, 154, 154}
\newcommand{\chartcolor}[1]{\textcolor{PrimaryBlue}{#1}}
\newtcolorbox{prompt}[1][]{
    breakable,
    colback=PrimaryBlue!5!,
    colframe=PrimaryBlue!0,
    colbacktitle=PrimaryBlue!70,
    rounded corners,
    sharp corners=northeast, 
    sharp corners=southwest,
    floatplacement=floating,
    title=\centering\textsf{\small #1}
}
\newtcolorbox{leftprompt}[1][]{
    breakable,
    colback=PrimaryBlue!5!,
    colframe=PrimaryBlue!0,
    colbacktitle=PrimaryBlue!70,
    rounded corners,
    sharp corners=northeast, 
    sharp corners=southwest,
    floatplacement=floating,
    title=\raggedright\textsf{\small #1}
}
\newcolumntype{L}[1]{>{\raggedright\arraybackslash}p{#1}}

\begin{prompt}[Generator (first-pass generation)]
\textbf{\#\#\# Task Description} \\
You are an expert in time-series classification. \\
Your task is to classify a time-series image into the correct category. \\

\textbf{\#\#\# Instruction} \\
- Analyze the image features and make a judgment based on the provided knowledge bank rules and other provided context. \\
- You can use the inter-class features to eliminate some categories, and then use intra-class features for pattern matching. \\
- When there are contradictions among evidence, follow the distinguishing insights. Recheck your thinking trajectory before output. \\

\textbf{\#\#\# Dataset Information}

\textcolor{BrownRed}{\{ dataset information including dataset name, description, number of classes, sequence length, etc. \}} \\

\textbf{\#\#\# Query Sample Information}

\textcolor{BrownRed}{\{ statistical information of query sample, including min and max value, mean value, variance, etc. \}} \\

\textbf{\#\#\# Knowledge Bank}

\textcolor{BrownRed}{\{ knowledge bank \}} \\

\textbf{\#\#\# Output Format}

Output your response strictly in JSON format with the following keys:

- reasoning: your thinking trajectory.

- bullet\_ids: list of the knowledge bank bullet points used in your reasoning.

- final\_answer: your concise final predicted label. \\

\textbf{\#\#\# Query Sample:}

Below is the Query Image to be classified:

\textcolor{BrownRed}{\{ visualized time series \}} \\
\end{prompt}

\begin{prompt}[Generator (second-pass generation)]
\textbf{\#\#\# Task Description} \\
You are an expert in time-series classification. \\
You are given a reasoning trace generated by another predictor, which should be treated as suspicious evidence, not ground truth. \\
Your task is to re-check the classification carefully using the playbook and the query image. \\

\textbf{\#\#\# Instruction} \\
- Act like a second-pass reviewer: explicitly look for errors, overconfidence, and missing evidence in the prior reasoning before deciding.\\
- If the first-pass reasoning is weak, revise it. If it is strong, you may keep the same answer, but only after independent verification. \\

\textbf{\#\#\# Dataset Information} \\
\textcolor{BrownRed}{\{ dataset information \}} \\

\textbf{\#\#\# Query Sample Information}\\
\textcolor{BrownRed}{\{ statistical information \}} \\

\textbf{\#\#\# Knowledge Bank}\\
\textcolor{BrownRed}{\{ knowledge bank \}} \\

\textbf{\#\#\# First-pass Generator Output}\\
Please re-evaluate the query image provided at the end and classify it.\\
Use the prior reasoning only as a doubtful clue, and correct it whenever necessary. \\
\textcolor{BrownRed}{\{ first-pass generator output \}} \\

\textbf{\#\#\# Output Format}

Output your response strictly in JSON format with the following keys:

- reasoning: your thinking trajectory.

- bullet\_ids: list of the knowledge bank bullet points used in your reasoning.

- final\_answer: your concise final predicted label. \\

\textbf{\#\#\# Query Sample:}

Below is the Query Image to be classified:

\textcolor{BrownRed}{\{ visualized time series \}} \\
\end{prompt}

\begin{prompt}[Reflector (reflect on correct prediction)]
\textbf{\#\#\# Task Description} \\
You are an expert in time series classification and pattern recognition. \\
Your task is to analyze a correct classification conducted by the previous predictor and summarize the successful experience. \\

\textbf{\#\#\# Instruction} \\
- Analyze the model's reasoning to identify the successful strategies and key insights. \\
- Evaluate the specific knowledge bank bullet points used by the model. Were they genuinely useful, neutral, or actually useless but the model succeeded anyway? \\

\textbf{\#\#\# Dataset Information} \\
\textcolor{BrownRed}{\{ dataset information \}} \\

\textbf{\#\#\# Query Sample Information} \\
\textcolor{BrownRed}{\{ statistical information of query sample \}} \\

\textbf{\#\#\# Knowledge Bank} \\
\textcolor{BrownRed}{\{ knowledge bank \}} \\

\textbf{\#\#\# Predictor Execution Details} \\
Please analyze this execution and provide your reflection based on the system instructions. \\
\textcolor{BrownRed}{\{ truth label of the query sample and output of generator \}} \\

\textbf{\#\#\# Output Format} \\
Output your response strictly in JSON format with the following keys: \\
- reasoning: your analysis of why the predictor succeeded \\
- key\_insight: the core successful strategy or experience that should be remembered. \\
- advice: the advice on how to improve the knowledge bank (empty if not available). \\
- bullet\_tags: your analysis about the bullet points usage of the predictor. Output in the form of a List[Dict], each dictionary should contain the following keys: (1) id, the index of the bullet point; (2) tag, indicating the role this bullet played in the reasoning process, which must be one of 'useful', 'useless', or 'neutral'; (3) reason, a brief explanation of your evaluation. \\

\textbf{\#\#\# Query Sample:} \\
Below is the Query Image to be classified by the predictor: \\
\textcolor{BrownRed}{\{ visualized time series \}} \\

\end{prompt}

\begin{prompt}[Reflector (reflect on incorrect prediction)]
\textbf{\#\#\# Task Description} \\
You are an expert in time series classification and pattern recognition.

Your task is to diagnose how the provided erroneous reasoning conducted by a predictor went wrong by analyzing the difference between the predicted label and the ground truth, then distill failure insight. \\

\textbf{\#\#\# Instruction} \\
- Carefully analyze the predictor's reasoning trace to identify where it went wrong. \\
- Identify specific conceptual errors, pattern recognition mistakes, or misapplied strategies. \\
- Provide actionable insights that could help the model avoid this mistake in the future. \\
- Focus on the root cause, not just surface-level errors. Be specific about what the model should do to avoid such mistake again. \\
- You will also receive several sample images from predicted label and the truth label. Identify the differences between these two categories and consider what factors led to incorrect classification in the previous reasoning. \\

\textbf{\#\#\# Dataset Information} \\
\textcolor{BrownRed}{\{ dataset information \}} \\

\textbf{\#\#\# Query Sample Information} \\
\textcolor{BrownRed}{\{ statistical information of query sample \}} \\

\textbf{\#\#\# Knowledge Bank} \\
\textcolor{BrownRed}{\{ knowledge bank \}} \\

\textbf{\#\#\# Predictor Execution Details} \\
Please analyze this execution and provide your reflection based on the system instructions. \\
\textcolor{BrownRed}{\{ truth label of the query sample and output of generator \}} \\

\textbf{\#\#\# Output Format} \\
Output your response strictly in JSON format with the following keys:\\
- error\_identification: what specifically went wrong in the original reasoning?\\
- root\_cause\_analysis: e.g., why did this error occur? What concept was misunderstood? What features are missing or overemphasized in the knowledge bank? What is the confusion point on the decision boundary?\\
- correct\_reasoning: the correct evidence-based reasoning path \\
- key\_insight: e.g., what strategy or principle should be remembered to avoid this error? \\
- bullet\_tags: your analysis of the usage of the bullet points in the predictor. Output in the form of a List[Dict], each dictionary should contain the following keys: (1) id, the index of the bullet point; (2) tag, indicating the role this bullet played in the reasoning process, which must be one of 'useful', 'useless', or 'neutral'; (3) reason, a brief explanation of your evaluation. \\

\textbf{\#\#\# Query Sample:} \\
Below is the Query Image to be classified by the predictor: \\
\textcolor{BrownRed}{\{ visualized time series \}} \\

\textbf{\#\#\# Sample in Predicted Label:} \\
\textcolor{BrownRed}{\{ visualized time series \}} \\

\textbf{\#\#\# Sample in Truth Label:} \\
\textcolor{BrownRed}{\{ visualized time series \}} \\

\end{prompt}

\begin{prompt}[Modifier]
\textbf{\#\#\# Task Description} \\
You are a master curator of a time series classification knowledge bank. \\ 
The knowledge bank you created will be used to help answer similar classification questions. \\
Your job is to identify what new insights should be added to an existing knowledge bank, what existing rules should be modified, or what to delete based on a reflection from a previous attempt.

\textbf{\#\#\# Instruction} \\
- Review the existing knowledge bank and the reflection from the previous reasoning.\\
- Identify ONLY the NEW insights, strategies, or mistakes that are MISSING from the current knowledge bank, or identify rules that need modification/deletion.\\
- Be concise and specific; each bullet point must be actionable. \\
- Return an empty list in the operations field if there is no need to modify the current knowledge bank.\\

\textbf{\#\#\# Dataset Information} \\
\textcolor{BrownRed}{\{ dataset information \}} \\

\textbf{\#\#\# Query Sample Information} \\
\textcolor{BrownRed}{\{ statistical information of query sample \}} \\

\textbf{\#\#\# Knowledge Bank} \\
\textcolor{BrownRed}{\{ knowledge bank \}} \\

\textbf{\#\#\# Predictor Execution Details} \\
\textcolor{BrownRed}{\{ truth label of the query sample and output of generator \}} \\

\textbf{\#\#\# Previous Reflection} \\
\textcolor{BrownRed}{\{ output of reflector \}} \\

\textbf{\#\#\# Output Format} \\
Output your response strictly in JSON format with the following keys:\\
- reasoning: your thinking trajectory. \\
- operations: operations you plan to perform on the knowledge bank. \\
\\
Available Operations: \\
1. ADD: Create new bullet points with fresh IDs. \\
- section: the section to add the new bullet to. \\
- content: the new content of the bullet. \\
2. MODIFY: Update an existing bullet point. \\
- target\_id: the exact ID of the bullet to modify. \\
- content: the fully updated content of the bullet. \\
3. DELETE: Remove an existing bullet point. \\
- target\_id: index of the bullet point you want to remove. \\

\textbf{\#\#\# Query Sample:} \\
Below is the Query Image to be classified by the predictor: \\
\textcolor{BrownRed}{\{ visualized time series \}} \\
\end{prompt}

\section{Detailed Results \& Extended Experiments}
\label{sec:more_analysis}

\subsection{Detailed Results of Increasing-Shot Comparison}
Figure~\ref{fig:comparison} in the main paper visualizes the overall trend of the comparison between \method{} under $k$-shot setting and baselines under increasing $k$-shot setting. Table~\ref{tab:fewshot_detailed_results} reports the detailed numerical results to better clarify the performance of \method{}.

\begin{table*}[t]
    \centering
    \caption{Detailed results of comparison with baselines under increasing-shot}
    \vspace{-1em}
    \label{tab:fewshot_detailed_results}
    \resizebox{\textwidth}{!}{
    \begin{tabular}{l*{21}{c}}
        \toprule
        \multirow{2}{*}{Dataset}
        & \multicolumn{1}{c}{Ours}
        & \multicolumn{5}{c}{TimesNet}
        & \multicolumn{5}{c}{PatchTST}
        & \multicolumn{5}{c}{GPT4TS}
        & \multicolumn{5}{c}{MOMENT} \\
        
        \cmidrule(lr){2-2}
        \cmidrule(lr){3-7}
        \cmidrule(lr){8-12}
        \cmidrule(lr){13-17}
        \cmidrule(lr){18-22}
        
        & $k$
        & $k$ & $2k$ & $3k$ & $4k$ & Full-shot
        & $k$ & $2k$ & $3k$ & $4k$ & Full-shot
        & $k$ & $2k$ & $3k$ & $4k$ & Full-shot
        & $k$ & $2k$ & $3k$ & $4k$ & Full-shot \\
        \midrule
        
        DodgerLoopGame
        & \textbf{0.884}
        & 0.681 & 0.761 & 0.761 & 0.833 & 0.833
        & 0.522 & 0.572 & 0.739 & 0.638 & 0.638
        & 0.529 & 0.565 & 0.695 & 0.699 & 0.699
        & 0.710 & 0.717 & 0.818 & 0.804 & 0.804 \\
        
        DistalPhalanxTW
        & 0.576
        & 0.467 & 0.518 & 0.573 & 0.612 & 0.705
        & 0.144 & 0.468 & 0.417 & 0.511 & 0.626
        & 0.612 & 0.612 & 0.604 & 0.604 & 0.705
        & 0.511 & 0.554 & 0.547 & 0.576 & \textbf{0.727} \\
        
        Fish
        & 0.771
        & 0.506 & 0.571 & 0.691 & 0.726 & 0.829
        & 0.183 & 0.331 & 0.320 & 0.480 & 0.491
        & 0.286 & 0.291 & 0.371 & 0.377 & 0.440
        & 0.709 & 0.726 & 0.857 & 0.869 & \textbf{0.914} \\
        
        GesturePebbleZ1
        & \textbf{0.785}
        & 0.302 & 0.343 & 0.302 & 0.349 & 0.390
        & 0.204 & 0.209 & 0.180 & 0.308 & 0.308
        & 0.168 & 0.157 & 0.192 & 0.221 & 0.250
        & 0.198 & 0.302 & 0.343 & 0.308 & 0.401 \\
        
        \bottomrule
    \end{tabular}
    }
\end{table*}

\subsection{Comparison with Extended Baselines}
We add the following four new baselines in comparison: \textbf{DTW} \citep{berndt1994dtw}: a classical few-shot machine learning method; \textbf{COSCO} \citep{barreda2024cosco} and its earlier baseline \textbf{TapNet} \citep{zhang2020tapnet}: two deep learning-based methods \textit{designed with few-shot-tailored adaptation}, and \textbf{TimeOmni-VL} \citep{guan2026timeomnivl}: a time-series VLM, which serves as a representative baseline with \textit{cross-modal alignment capability}. For DTW and TapNet, we use the implementation from AEON \citep{middlehurst2024aeon}. For COSCO, we train it on few-shot train set with default settings as detailed in the paper, i.e., $lr=0.01$ and 100 epochs. For TimeOmni-VL, we use it without fine-tuning, provide the same prompt and knowledge bank as \method{} to supply task background, while keeping all generation hyperparameters at default values.

Table~\ref{tab:extended_baseline} shows that \method{} consistently outperforms both few-shot-tailored and cross-modal-aligned baselines. This result supports that the performance gain mainly comes from the proposed agentic reasoning framework. In particular, the comparison with COSCO shows the advantage of \method{} over dedicated few-shot methods. The comparison with TimeOmni-VL further indicates that directly applying a TS-aligned VLM remains insufficient without structured reflective refinement and knowledge evolution.

\begin{table}[H]
    \centering
    \vspace{-1.5em}
    \caption{Accuracy Comparison with extended baselines}
    \label{tab:extended_baseline}
    \vspace{-0.15in}
    \small
    \renewcommand{\arraystretch}{0.9}
    \setlength{\tabcolsep}{1.6pt}
    \begin{tabular}{l|ccccc}
        \toprule
        Dataset & \method{} & DTW & COSCO & TapNet & TimeOmni-VL \\
        \midrule
        DodgerLoopGame & \textbf{0.884} & 0.662 & 0.567 & 0.642 & 0.541 \\
        DiatomSizeReduction & \textbf{0.971} & 0.906 & 0.952 & 0.590 & 0.265 \\
        DistalPhalanxTW & \textbf{0.576} & 0.542 & 0.528 & 0.515 & 0.212 \\
        Fish & \textbf{0.771} & 0.564 & 0.661 & 0.496 & 0.145 \\
        GesturePebbleZ1 & \textbf{0.785} & 0.648 & 0.732 & 0.519 & 0.165 \\
        \bottomrule
    \end{tabular}
\end{table}

\subsection{Analysis on Test-Time Computation}
To isolate the effect of test-time computation, we conduct a new experiment to compare our method with a test-time scaling variant based on $m$-voting self-consistency. The rationale for this new comparison is that it provides a scenario of \textit{spending comparable inference budget without agentic reasoning}. Specifically, for a dataset with $c$ classes, we perform $m \times c$ inference calls and take the majority vote as the final prediction. The voting variant uses the same prompt as Generator agent and warm-up knowledge bank.

\begin{figure}[h]
    \centering
    \vspace{-0.4em}
    \includegraphics[width=\columnwidth, height=2cm]{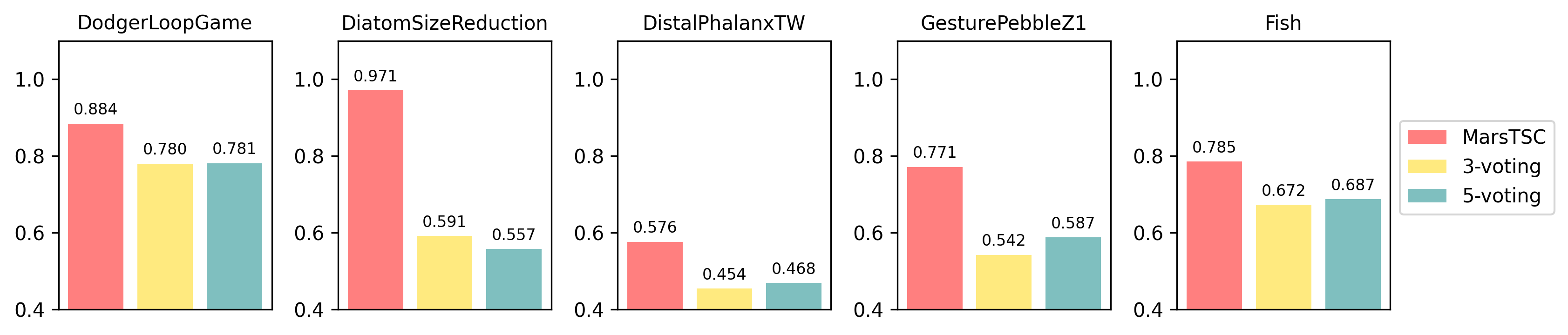} 
    \vspace{-2em}
    \caption{Comparison with test-time scaling method}
    \vspace{-0.9em}
\label{fig:rebuttal_test_time_computation} 
\end{figure}

As shown in Figure~\ref{fig:rebuttal_test_time_computation}, increased computation budget can improve performance, but they still underperform the full \method{} pipeline. This suggests that larger test-time computation alone cannot explain the observed gains. The key advantage of \method{} lies in how the computation is structured: instead of simply aggregating multiple predictions, \method{} uses insight distillation and cautious deferred updates to convert reasoning outcomes into reusable evolving knowledge.

\subsection{Analysis of Few-shot Train Samples} 
\begin{figure}[h]
    \centering 
    \includegraphics[width=0.45\textwidth]{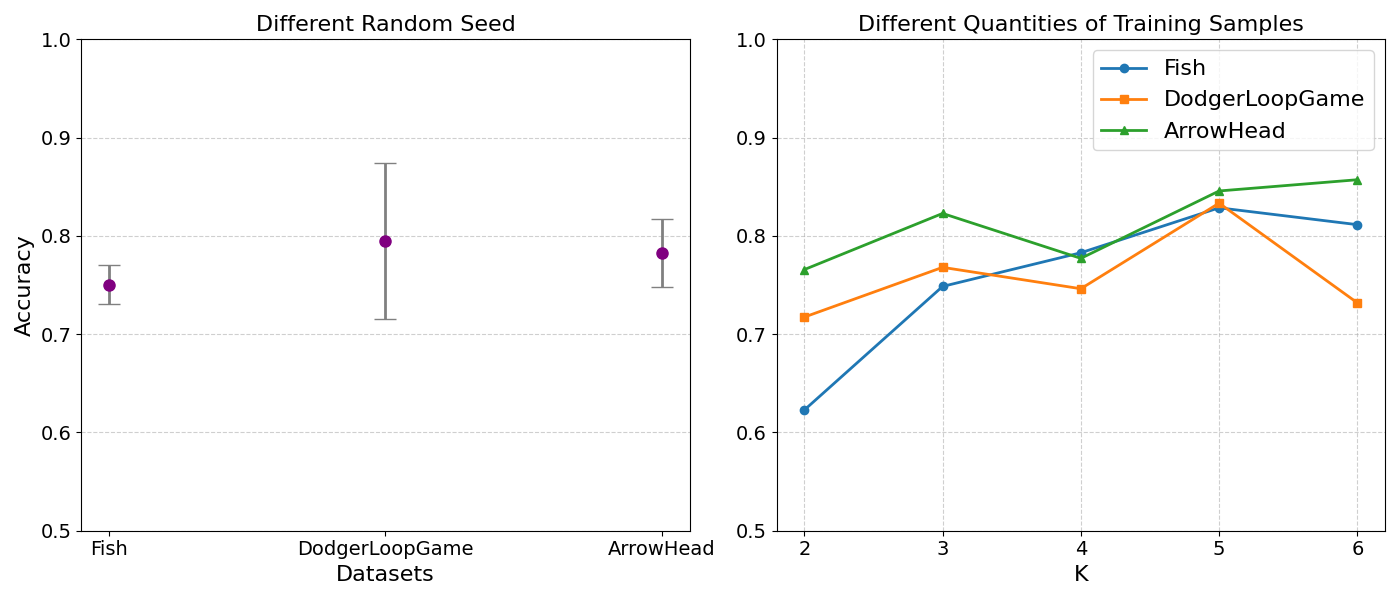} 
    \vspace{-1.25em}
    \caption{Left: Comparison of mean accuracy and standard deviation with different random seeds; Right: Accuracy of three datasets across different $k$-shot settings.}
\label{fig:shot} 
\end{figure}

To further evaluate the robustness of \method{} under few-shot sampling randomness and varying levels of data scarcity, we conduct extended experiments on three datasets with different $k$-shot settings ($k \in \{2, 3, 4, 5, 6\}$) and random seeds. Figure~\ref{fig:shot} shows that the error bars remain relatively small, indicating stable performance across different few-shot selections. In addition, the mean accuracy generally improves as the number of shots increases, demonstrating that the dynamic knowledge bank can progressively incorporate new discriminative patterns in an effective and stable manner. While a minor fluctuation appears on DodgerLoopGame at the 6-shot setting, likely caused by anomalous samples, the overall trend confirms the strong reliability and generalization ability of \method{} across diverse few-shot conditions.

\subsection{Analysis of Knowledge Bank Size} 
\begin{figure}[t]
    \centering 
    \includegraphics[width=0.45\textwidth]{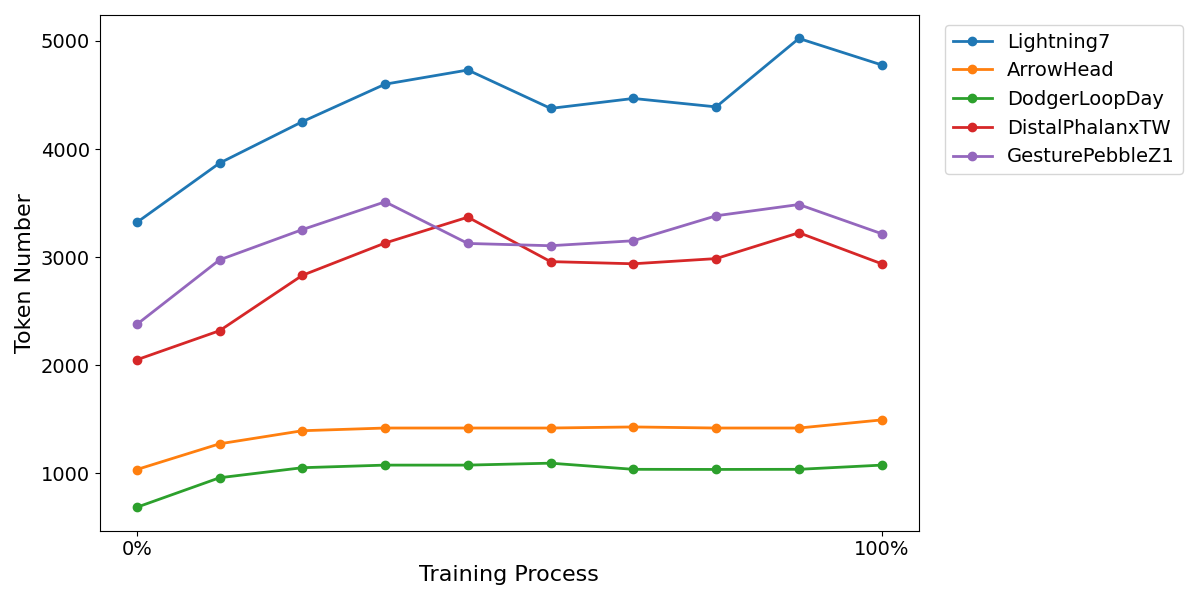} 
    \vspace{-1.25em}
    \caption{Token count of the knowledge bank over the course of training.}
\label{fig:token} 
\end{figure}

To further analyze the behavior of the evolving knowledge bank, we monitor its token count during training across different time-series datasets. Figure~\ref{fig:token} shows that the token count increases only in the early stage and then gradually stabilizes, rather than expanding without bound. This suggests that the proposed bullet-level update mechanism is able to selectively preserve informative knowledge while filtering out redundant or low-value content. Such a design mitigates the risk of context degradation or collapse caused by repeatedly rewriting the entire knowledge bank, thereby allowing agents to operate with a compact, discriminative, and computationally efficient context during iterative reasoning.

\subsection{Ablation on Key Stages}
\label{ab_on_key_stages}
To understand the contribution of each stage, we compare the full pipeline with two reduced variants: \textit{w/o test update} disables test-time update, and \textit{w/o train refine \& test update}, disables both train-time refinement and test-time update, which performs testing solely based on the initial statistical features and visual descriptions. Notably, \textit{w/o train refine \& test update} can be regarded as a setting that mainly reflects the classification ability of the VLM itself, since no agentic operation is involved.

As shown in Table~\ref{tab:ablation_key_stage}, \textit{w/o test update} consistently outperforms \textit{w/o both} on all datasets, indicating that statistical features and visual descriptions alone are insufficient for effective classification, highlighting the importance of the agent reasoning, whose role is not to directly make the final prediction but to accumulate successful experience, extract corrective feedback from failures, and refine the inference process accordingly.

Moreover, in most cases, the full pipeline further outperforms \textit{w/o test update}, suggesting that the proposed test-time update mechanism can uncover discriminative patterns not fully covered by the few-shot samples, thereby alleviating the limited representativeness and generalization of few-shot supervision.

\begin{table}[t]
    \centering
    \caption{Ablation on key stages}
    \label{tab:ablation_key_stage}
    \vspace{-0.1in}
    \small
    \renewcommand{\arraystretch}{0.9}
    \setlength{\tabcolsep}{3.5pt}
    \begin{tabular}{l|ccc}
        \toprule
        Dataset & Full pipeline & w/o test update & \makecell{w/o train refine\\\& test update} \\
        \midrule
        ArrowHead & \textbf{0.766} & 0.691 & 0.543 \\
        DodgerLoopGame & \textbf{0.884} & 0.812 & 0.797 \\
        ECG200 & \textbf{0.790} & 0.770 & 0.640 \\
        GesturePebbleZ1 & \textbf{0.785} & 0.756 & 0.744 \\
        Lightning7 &\textbf{ 0.822} & 0.795 & 0.740 \\
        UMD & \textbf{0.993} & 0.993 & 0.924 \\
        Worms & \textbf{0.481} & 0.464 & 0.442 \\
        DiatomSizeReduction & \textbf{0.971} & 0.971 & 0.771 \\
        DistalPhalanxTW & \textbf{0.576} & 0.561 & 0.510 \\
        Fish & \textbf{0.771} & 0.726 & 0.600 \\
        RacketSports & \textbf{0.526} & 0.526 & 0.454 \\
        SelfRegulationSCP1 & 0.697 & \textbf{0.717} & 0.519 \\
        \bottomrule
    \end{tabular}
\end{table}

\subsection{Analysis of Base VLM Sensitivity}
\label{ab_on_vlm}
To further investigate the robustness of our method across different foundation models, we additionally evaluate \method{} with four other mainstream VLMs, including GPT-5~\citep{openai_gpt5}, GPT-5.4 mini~\citep{openai_gpt54mini}, Qwen-3.5-397B-A17B~\citep{qwen35_397b}, and Kimi-K2.5~\citep{kimi_k25}.

\begin{figure}[h]
    \centering 
    \includegraphics[width=0.5\textwidth]{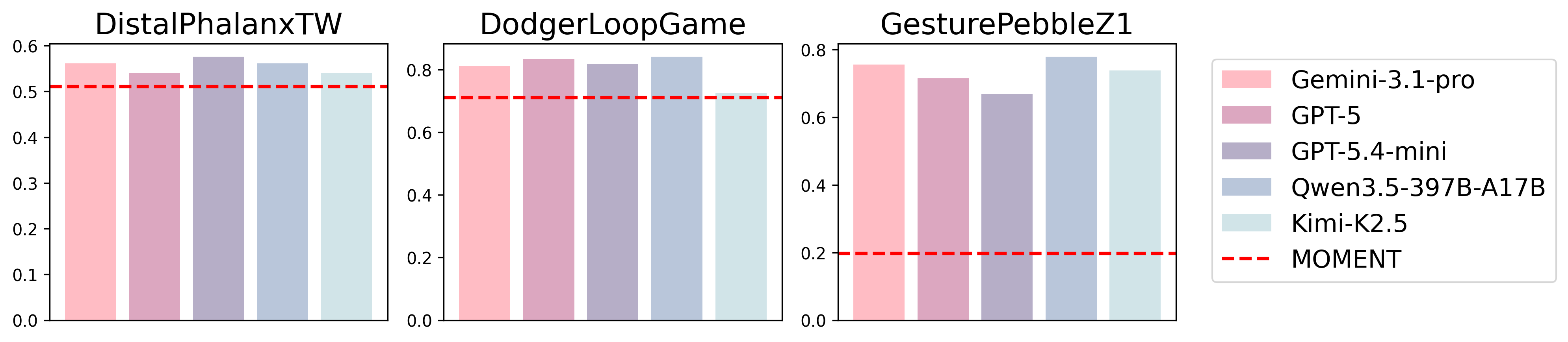} 
    \caption{Classification accuracy with different base VLMs}
\label{fig:ab_on_model} 
\end{figure}

As illustrated in Figure~\ref{fig:ab_on_model}, the overall performance of our method varies slightly with the capability of the base VLM, which is expected given the different reasoning and multimodal understanding abilities of these models. Nevertheless, in most cases, our method still outperforms most baselines shown in Table~\ref{tab:result_few_shot}. This suggests that our method can be effectively transferred across different VLM backbones while retaining strong performance.

\subsection{More Case Studies}
\label{sec:more_case_studies}
\begin{figure}[H]
    \centering 
    \includegraphics[width=0.95\linewidth]{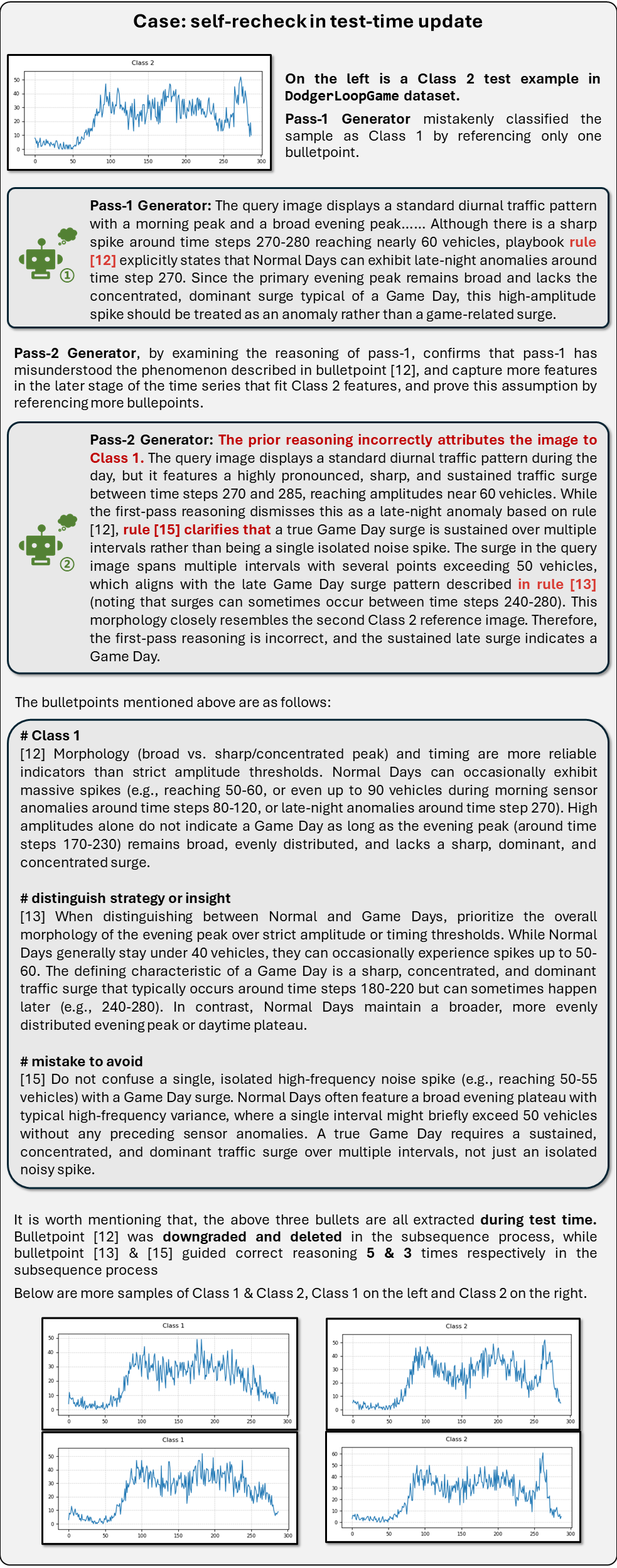} 
    \caption{A case about self-recheck in test time. By using prior reasoning as suspicious reference, generator successfully identified the original mistake and corrected it.}
\label{fig:case_3_detailed} 
\end{figure}

Figure~\ref{fig:case_3_detailed} presents a successful example of test-time self-recheck via two-pass generation, where the model revisits its previous reasoning, corrects the initial mistake, and ultimately arrives at the correct answer. This case suggests that the test time design can effectively reduce the risk of error-induced knowledge contamination while further expanding the coverage and adaptability of the knowledge memory.

\begin{figure}[H]
    \centering
    \includegraphics[width=0.95\linewidth]{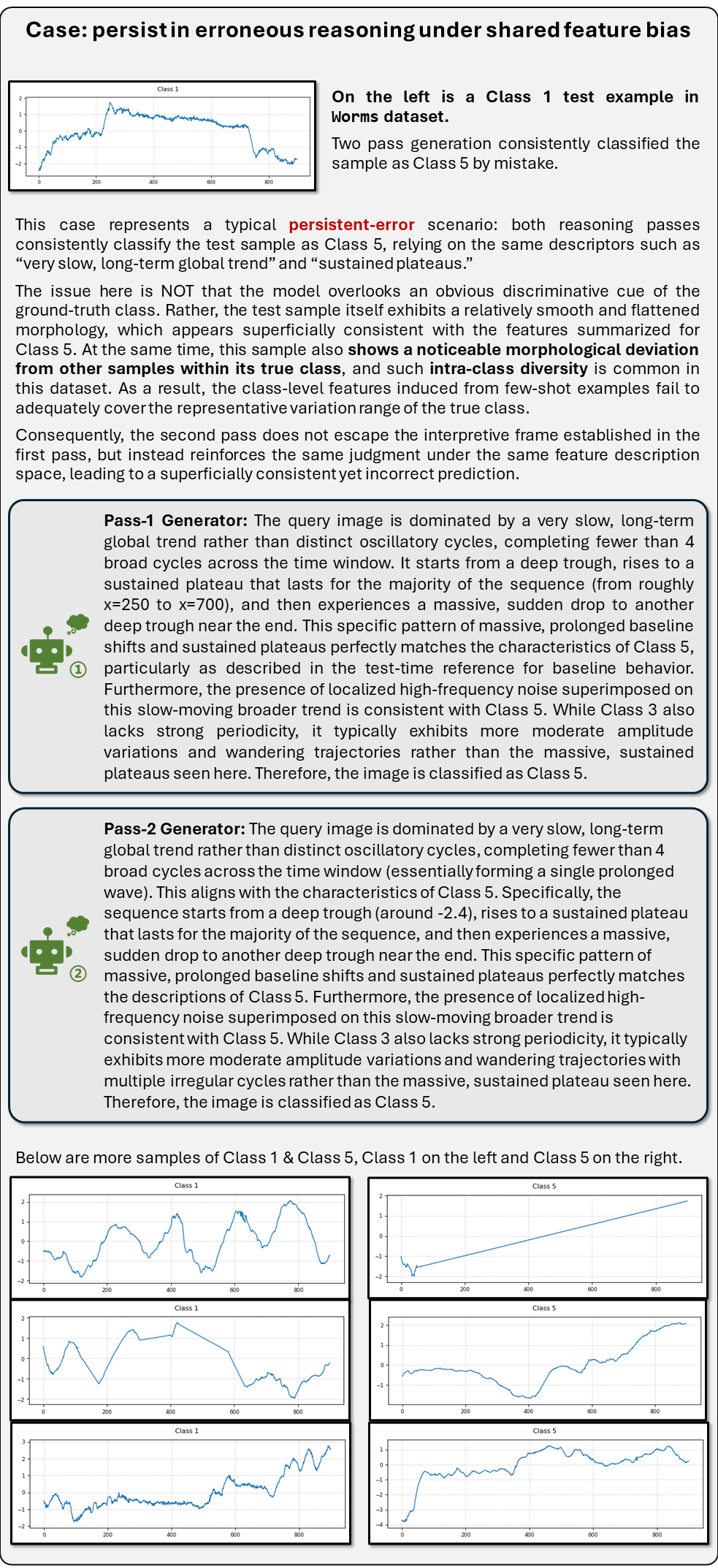} 
    \caption{A case about self-recheck in test time. Pass-2 generation failed to correct the answer due to insufficient coverage of pattern diversity in the few-shot examples.}
\label{fig:case_4} 
\end{figure}

\begin{figure}[H]
    \centering 
    \includegraphics[width=0.95\linewidth]{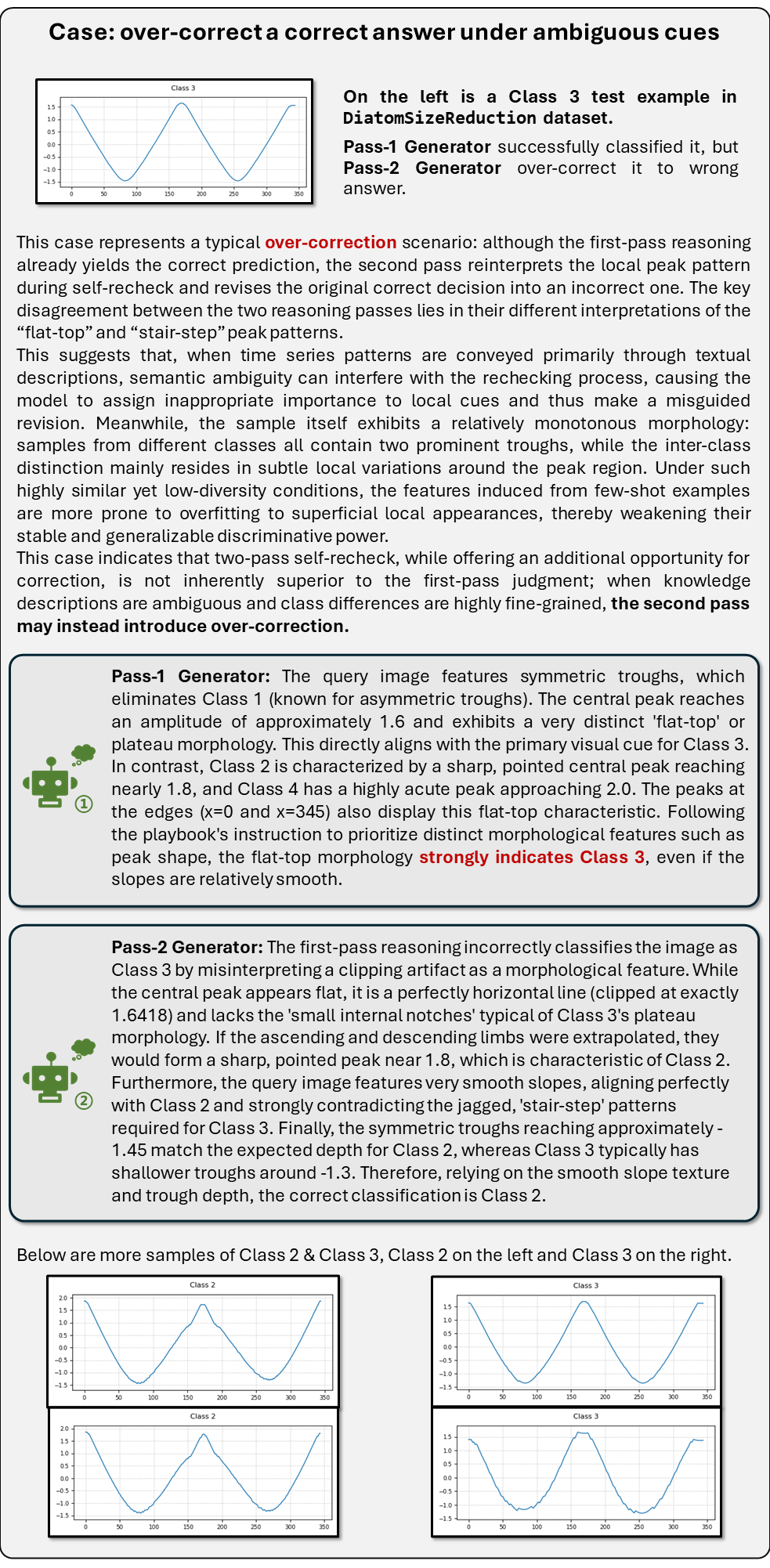} 
    \caption{A case about self-recheck in test time. Pass-2 reasoning revises a correct initial prediction due to an ambiguous description in the knowledge bank.}
\label{fig:case_5} 
\end{figure}

We further present two failure cases to clarify the effective boundary of the proposed two-pass design. In the first case, as shown in Figure~\ref{fig:case_4}, both passes produce the same incorrect prediction, indicating that the second pass does not always constitute a genuinely independent re-evaluation. Instead, it may remain anchored to the initial hypothesis and further reinforce a self-consistent but incorrect reasoning. Figure~\ref{fig:case_5} provides the second case, in which the first pass gives the correct answer, whereas the second pass revises it to an incorrect one. This suggests that self-recheck may occasionally introduce over-correction, where additional reasoning shifts attention toward misleading or non-discriminative cues rather than improving the original judgment.

Taken together, these cases show that two-pass generation should be understood as a risk-reduction mechanism rather than a guarantee of correction. While it can often correct initial mistakes, its effectiveness depends on whether the second pass can both escape first-pass anchoring and rely on sufficient discriminative knowledge. This observation also motivates our conservative deferred knowledge update strategy during test time, which avoids directly treating second-pass outputs as fully trustworthy knowledge.

\section{Future Directions}
\label{sec:future_directions}

Several promising directions warrant further investigation. First, a unified time-series reasoning framework could extend the proposed \method{} beyond classification to forecasting, anomaly detection, imputation, and temporal question answering. Such a framework would require task-aware knowledge structures and reasoning policies that capture both shared temporal characteristics and task-specific decision objectives. Second, richer visual representations, such as spectrograms or recurrence plots, deserve systematic exploration. These representations may help VLM agents better capture frequency-domain patterns that are difficult to identify from line plots alone. Third, integrating more capable agents with tool-use and long-term memory represents another important direction. External tools can support evidence verification, while persistent memory can retain reusable knowledge across datasets and tasks, improving the reliability and transferbility of \method{}.

\newpage

\end{document}